\documentclass[review]{elsarticle}
\usepackage{amsmath}
\usepackage{lineno,hyperref}
\usepackage{setspace}

\usepackage{geometry}
\usepackage{multirow}
\usepackage{amssymb}
\usepackage{wasysym}

\usepackage{setspace}
\usepackage{changes}
\usepackage{makecell}
\usepackage{threeparttable} 
\definechangesauthor[name={Shuer}, color=orange]{S.}

\biboptions{authoryear}

\begin{document}

\begin{frontmatter}

\title{A review of individual tree crown detection and delineation from optical remote sensing images}


\author[SYS1]{Juepeng Zheng \fnref{myfootnote}}
\author[SYS1,XA]{Shuai Yuan \fnref{myfootnote}} 
\author[SYS2]{Weijia Li\corref{ca1}}
\author[THU,WX,XA]{Haohuan Fu\corref{ca1}}
\author[THU]{Le Yu}

\address[SYS1]{School of Artificial Intelligence, Sun Yet-Sen University, Zhuhai, China}
\address[THU]{Ministry of Education Key Laboratory for Earth System Modeling, Department of Earth System Science, Tsinghua University, Beijing, China}
\address[XA]{Tsinghua University (Department of Earth System Science)- Xi'an Institute of Surveying and Mapping Joint Research Center for Next-Generation Smart Mapping, Beijing, China}
\address[WX]{National Supercomputing Center in Wuxi, Wuxi, China.}
\address[SYS2]{School of Geospatial Engineering and Science, Sun Yat-Sen University, Zhuhai, China}

\cortext[ca1]{Corresponding author: haohuan@tsinghua.edu.cn}
\fntext[myfootnote]{These authors contributed equally to this work.}


\begin{abstract}
%Since the advances in optical remote sensing sensors during the last two decades have led to the production of very high spatial resolution multispectral images, there is a great potential for achieving high accuracy forest inventory and analysis automatically and cost-efficiently. 
Powered by the advances of optical remote sensing sensors, the production of very high spatial resolution multispectral images provides great potential for achieving cost-efficient and high-accuracy forest inventory and analysis in an automated way. Lots of studies that aim at providing an inventory to the level of each individual tree have generated a variety of methods for Individual Tree Crown Detection and Delineation (ITCD). This review covers ITCD methods for detecting and delineating individual tree crowns, and systematically reviews the past and present of ITCD-related researches applied to the optical remote sensing images. With the goal to provide a clear knowledge map of existing ITCD efforts, we conduct a comprehensive review of recent ITCD papers to build a meta-data analysis, including the algorithm, the study site, the tree species, the sensor type, the evaluation method, etc. We categorize the reviewed methods into three classes: (1) traditional image processing methods (such as local maximum filtering, image segmentation, etc.); (2) traditional machine learning methods (such as random forest, decision tree, etc.); and (3) deep learning based methods. With the deep learning-oriented approaches contributing a majority of the papers, we further discuss the deep learning-based methods as semantic segmentation and object detection methods. 
%Second, we summarize existing ITCD methods and categorize them into three classes, including traditional image processing based ITCD methods, traditional machine learning based ITCD methods and deep learning based ITCD methods, while each of them has advantages and disadvantages. We further categorize deep learning based ITCD methods into CNN classification based, semantic segmentation based and object detection based ITCD methods.  
In addition, we discuss four ITCD-related issues to further comprehend the ITCD domain using optical remote sensing data, such as comparisons between multi-sensor based data and optical data in ITCD domain, comparisons among different algorithms and different ITCD tasks, etc. Finally, this review proposes some ITCD-related applications and a few exciting prospects and potential hot topics in future ITCD research. %We hope this review help researchers who involved in ITCD domain to keep track of its development and tendency.

\end{abstract}


\end{frontmatter}

%\linenumbers

\section{Introduction}

Trees contribute extensively to ecology, environmental, economic and society domains in both global and local regions \citep{hansen2013high}. Forest ecosystems contribute significantly to global biogeochemical cycles, harbor a large proportion of biodiversity, and provide countless ecosystem services, including carbon sequestration, timber stocks and water quality control \citep{bonan2008forests}.  Therefore, better forest management not only provides renewable resources for human activities, but also makes great contributions to ecological conservation and the global energy circle. On the other hand, urban trees play an essential role in the urban environment and have many significant meanings to urban residents, such as beauty, shade, cooling, and gas balances \citep{nowak2006air}.

It is necessary to assign a thorough forestry inventory for sustainable forest management, and a comprehensive urban tree survey for living environment improvement, which includes measuring and average tree height, tree diameter and tree ages, etc., and other parameters and information for individual trees, such as position, species and crown size, etc. \citep{yrttimaa2020performance}. In the traditional investigation period, surveyors regularly measure the parameters for individual trees by field surveys and zonal sampling. Since the 1960s, with the development of aerial photography, manual imagery interpretation has been widely applied in forest inventory. However, no matter the routine manpower field surveys, or the visual interpretation of aerial photographs, it costs a large amount of human labor, work time and expense. Fortunately, lots of commercial satellites start to collect earth observation data at a higher spatial resolution during the past two decades, which is able to capture ground objects measuring one square meter or less \citep{hanan2020satellites}. Compared to cost-prohibitive aerial or in-field surveys, satellite images are suitable to monitoring the dynamics of forest or urban trees, since it is an efficient way for large-scale and repeated surveys over time \citep{payne2021satellite}. In the meantime, along with the rapid progress of computer techniques, especially automatically detecting objects from digital image processing, it offers viable opportunities for automatic ITCD through high-resolution remote sensing images \citep{brandt2020unexpectedly}.

In this review, Individual Tree Crown Detection and Delineation (ITCD) includes detecting and delineating tree crowns. Individual tree crown detection is mainly oriented to the location of an individual tree, such as the center or the coordinates of four corners of the tree crown. Individual tree delineation primarily focuses on sketching the contour and shape of the tree crown or the area of tree crown canopy volume. These two can further contribute to specific tasks like individual tree counting which aims at one-by-one counting or estimating the number of trees \citep{crowther2015mapping}. ITCD can effectively estimate tree crown size, tree biomass, growing status, etc. In addition, large-scale ITCD results are conducive to the mapping of forest boundaries, tree density and tree species or acquisition to other forest parameters. To this end, the results of ITCD are basic and indispensable data in forest inventory, and with the assistance of plentiful remote sensing data and vigorous artificial intelligence techniques \citep{lecun2015deep}, automatic ITCD becomes a reality and plays an essential role in forest inventory and tree management.

\begin{table}[t]
    \centering
    \caption{\label{tab:SOTAreview}Summary of existing ITCD related reviews.} 
    \resizebox{\textwidth}{!}{
    \begin{tabular}{ccccccccc}
    \hline
       \multirow{2}*{Publicaitons} &  \multicolumn{3}{c}{Reviewed ITCD methods$^*$} &  \multicolumn{2}{c}{Reviewd ITCD topics} & \multicolumn{2}{c}{Reviewd ITCD tasks}& \multirow{2}*{Reviewed Data}    \\ 
       &  TIP & TML & DL & Detecting & Delineating & Counting  & Applications$^*$  \\\hline
       \citep{hyyppa2008review} & $\times$ & $\times$ & $\times$ & $\checkmark$ & $\checkmark$ & $\times$ & $\times$ & LiDAR data\\
       \citep{ke2011review} & $\checkmark$ & $\times$ & $\times$ & $\checkmark$ & $\checkmark$ & $\times$ & $\times$& Optical data   \\
       \citep{wulder2012lidar} & $\times$ & $\times$ & $\times$ & $\checkmark$ & $\checkmark$ & $\times$  & $\times$& LiDAR data\\
       \citep{zhen2016trends} & $\checkmark$ & $\times$ & $\times$ & $\checkmark$ & $\checkmark$ & $\times$  & $\times$& LiDAR data        \\
       %\citet{gomes2016detection} &  $\checkmark$ & $\times$ & $\times$ & $\checkmark$ & $\checkmark$ & $\times$ & $\times$ & Optical data\\
        \citep{yin2016assess} & $\times$ & $\times$ & $\times$ & $\checkmark$ & $\checkmark$ & $\times$ & $\times$ & LiDAR data \\
        Ours & $\checkmark$ & $\checkmark$ & $\checkmark$ & $\checkmark$ & $\checkmark$ & $\checkmark$  & $\checkmark$ &Optical data \\  \hline
    \end{tabular}
    }
        \begin{tablenotes}
	    \footnotesize 
        \item[1] $^*$TIP denotes Traditional image processing-based ITCD methods, TML denotes traditional machine learning-based ITCD methods, DL denotes deep learning-based ITCD methods.
        \item[2] $^*$Applications denote the ITCD-related applications such as tree species classification, health monitoring, and tree parameter estimation, etc.
        \end{tablenotes}
    
\end{table}

\begin{figure}[t]
    \centering
    \includegraphics[width=1.0\linewidth]{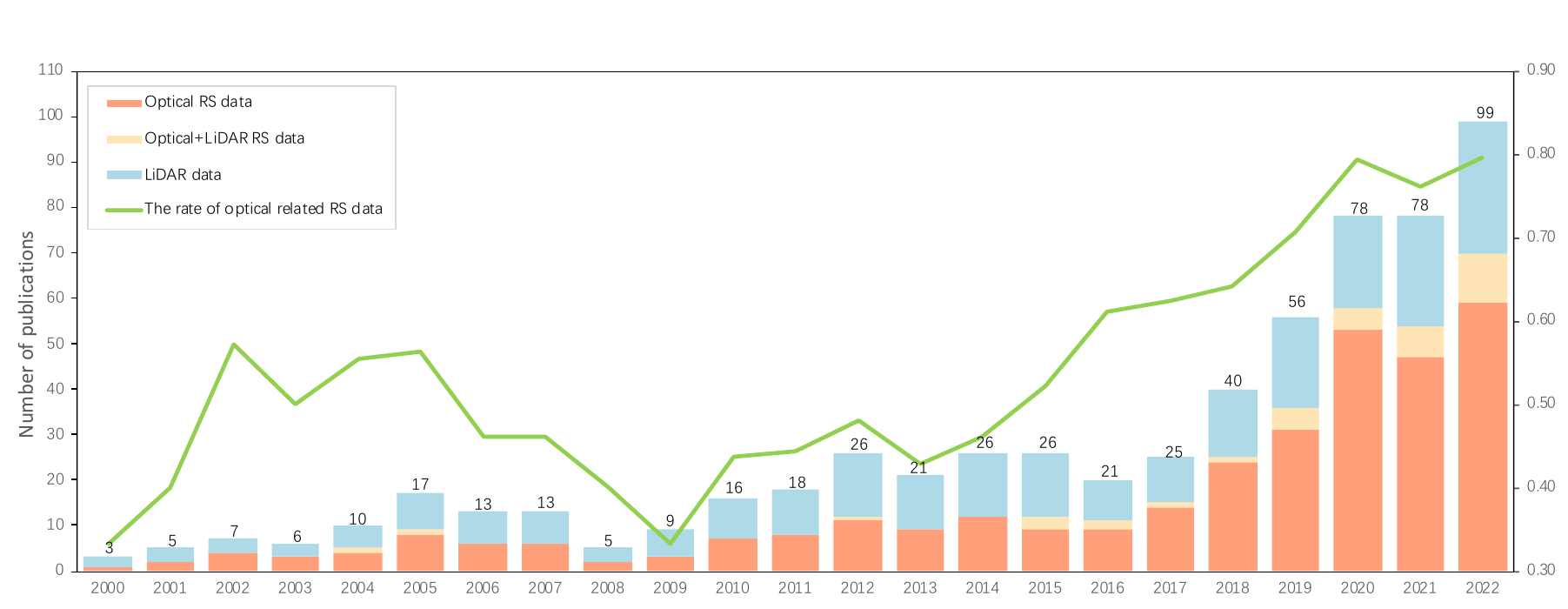}
    \vspace{-1em}
    \caption{The number of ITCD from LiDAR data and optical remote sensing data-related publications from 2000 to 2022. In past ten years, the rate of ITCD-related papers that used optical remote sensing data has steadily increased.} 
    \label{fig:optical_LiDAR}
\end{figure}

There is a variety of reviews about trees \citep{white2016remote}, including tree species classification \citep{fassnacht2016review,wang2019review,michalowska2021review}, fruit detection \citep{koirala2019deep} and yield estimation \citep{rashid2021comprehensive}. Most of them emphasize Light Detection And Ranging (LiDAR) \citep{michalowska2021review} or thermal imaging \citep{ahmed2019detection}. Some surveys only review one specific tree species, such as oil palm \citep{chong2017review}. 
Up to now, there are only five ITCD-related reviews, as listed in Tab. \ref{tab:SOTAreview}. \citet{hyyppa2008review} and \citet{wulder2012lidar}  only focus on LiDAR data and vertically distributed forest attributes estimation.
\citet{yin2016assess} reviews the available techniques for evaluating detected individual tree locations and crown delineation maps using remote sensing data. They mainly discussed ITCD assessment rather than ITCD methods. Recently, \citet{zhen2016trends} conducts a comprehensive survey for two of ITCD topics (detecting and delineating) using traditional image processing-based ITCD methods while they only focus on LiDAR data. \citet{ke2011review} reviews ITCD methods using passive remote sensing imagery, which only conducts a survey on traditional image processing-based ITCD methods. They categorize and evaluate algorithms for automatic tree crown detection (such as template matching,  scale analysis, image binarization and local maximum filtering) and delineation (such as watershed segmentation, region-growing and valley-following). However, it is so far the only comprehensive ITCD review that was published ten years ago.
%\added{\citet{gomes2016detection} also offers a similar review on traditional image processing based ITCD methods.} 
%\citet{gomes2016detection} also offer a similar review on traditional image processing based ITCD methods. \citet{hanapi2019review} classify ITCD methods into four types, including image processing-based method, machine learning-based method, point cloud-based method and deep learning-based method. Our taxonomic is quite similar to \citet{hanapi2019review}, however, they only review 60 ITCD related publications and have not conducted a thorough mate-analysis and in-depth discussion on this topic. In addition, we further classify deep learning based ITCD methods into three categories: CNN classification based ITCD algorithm, semantic segmentation based ITCD algorithm and object detection based ITCD algorithm. 
According to Tab. \ref{tab:SOTAreview}, existing ITCD reviews can not completely comprise of ITCD methods proposed by recent years, especially under the continuously rapid development of machine learning and deep learning algorithms (see Fig. \ref{fig:total}). Furthermore, various important ITCD-related tasks and applications (e.g., counting the number of trees, health monitoring, parameter estimation, etc.) are paid rare attention in existing reviews. On the other hand, as shown in Fig. \ref{fig:optical_LiDAR}, we can observe that the optical data has been more and more used in ITCD domain. In the last ten years, the rate of ITCD-related papers that used optical remote sensing data has steadily increased. Especially after 2020, most ITCD-related publications (nearly 80\%) adopt optical remote sensing data to achieve high-accuracy and large-scale tree crown detection.

% \added{12345}

% \deleted{12345}

% \replaced{12345}{123}

All in all, optical remote sensing data has become an essential avenue in the ITCD domain. It is necessary to summarize the characteristics and trends from ITCD-related research during the past ten years, helping readers comprehend the past, present and future of the ITCD domain. Notably, this review mainly surveys ITCD from optical remote sensing images, as well as combining optical remote sensing images and LiDAR data. Research that only adopts LiDAR data on ITCD is out of scope in this review.
The contributions of this review mainly include the following three points:

\begin{enumerate}
    \item We conduct a review of Individual Tree Crown Detection and Delineation (ITCD), including a meta-analysis of the literature, a thorough comparison of methodology, in-depth discussion, extensive related applications and potential prospects. This paper is the first systematic review of ITCD in recent ten years to the best of our knowledge.
    %This database served as the foundation for statistical analysis and can be available on \url{https://github.com/rs-dl/ITCDReview}.
    \item We discuss and analyze the pros and cons for all kinds of existing ITCD approaches in different scenarios from three aspects: traditional image processing approaches, traditional machine learning approaches, and deep learning approaches. We first review deep learning-based ITCD methods and conduct comparisons between general deep learning models and their applications in ITCD domain.
    \item We conduct in-depth discussion on pros and cons for different ITCD algorithms and tasks, and list extensive ITCD-related applications \& tasks, and envision promising future works in the ITCD domain. Furthermore, we point out that optical remote sensing data will become a key driver of future ITCD-related studies, including tree crowns detection, forestry inventory and other applications.  
    
\end{enumerate}

The remainder of this review is organized as follows. We present the meta-analysis of related literature in Sec. \ref{sec:meta}, such as data collection and quantitative analysis. Following that, we conduct a thorough review of the methodology of ITCD in Sec. \ref{sec:method}, including traditional image processing-based ITCD methods, traditional machine learning-based ITCD methods and deep learning-based ITCD methods. After that, we present assessment for the accuracy of ITCD methods in Sec. \ref{sec:assessment} In Sec. \ref{sec:diss}, we make an in-depth discussion on the comparison of different ITCD methods, characteristics of ITCD researches, assessment of the accuracy, etc., followed by extensive ITCD related applications, such as tree parameters, forest monitoring, etc. in Sec. \ref{sec:application}. We envision our promising prospects on ITCD domain in Sec. \ref{sec:pros}. Finally, we conclude this review in Sec. \ref{sec:concl}.

\section{Meta-analysis of related literature}
\label{sec:meta}

\begin{figure}[t]
    \centering
    \includegraphics[width=1.0\linewidth]{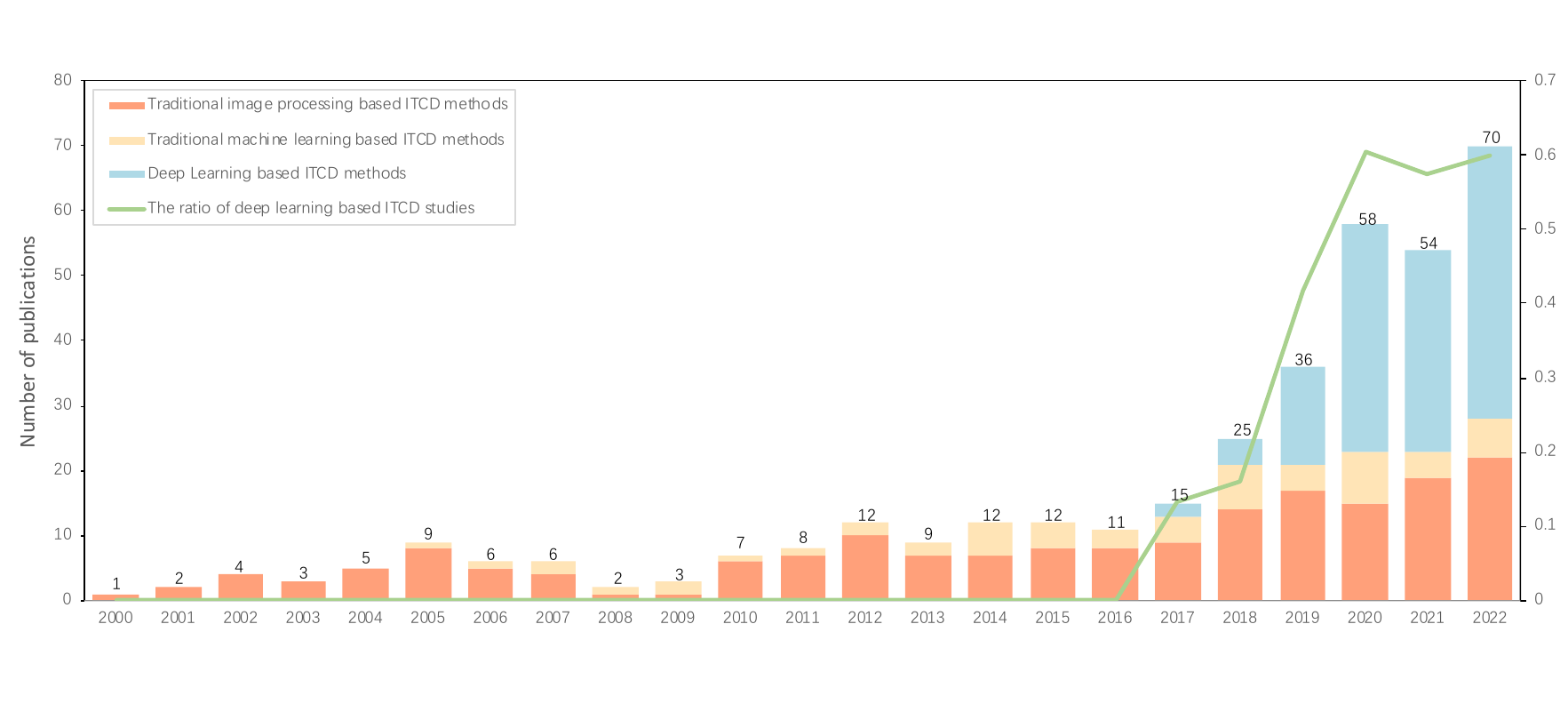}
    \vspace{-3em}
    \caption{The number of ITCD from optical remote sensing images related publications from 2000 to 2022. The literature survey was conducted in Web of Science, Scopus and Google Scholar. Since 2017, deep learning-based ITCD methods have become the most popular ITCD approach.}
    \label{fig:total}
\end{figure}

As shown in Fig. \ref{fig:total}, the number of ITCD-related papers using optical remote sensing data has exponentially increased since 2017, which is notoriously difficult to keep track of ITCD-related research for those who are involved in ITCD domain. To this end, it is essential to periodically conduct a review in order to summarize recently implemented ITCD methods, study areas, tree species and the types of optical remote sensing data. In this section, we conduct a meta-analysis regarding the ITCD domain to investigate these subjects. 

\begin{table}[t]
    \centering
    \caption{Database fields created in order to extract relevant information and conduct the meta-analysis.}
    \begin{tabular}{cccc}
    \hline
        ID & Field name & Description & Type  \\
        \hline
        1 & Title & Title of literature  & Text \\
        2 & Literature source & Title of source (journal) & Text \\
        3 & Year & Year of publication & Numeric \\
        4 & Research Institution & Country of research institution & Text \\
        5 & Case study site & Country of case study site & Text \\
        6 & Study area & Area of case study site & Numeric \\
        7 & Tree species & Species of tree in case study & Text \\
        8 & Sensor & Sensor type  & Text \\
        9 & Spatial resolution & Spatial resolution of data & Numeric \\
        10 & ITCD method & ITCD method implemented & Text \\
        11 & Evaluation & Evaluation of ITCD & Text \\
        \hline
    \end{tabular}
    \label{tab:database}
\end{table}

\begin{figure}[t]
    \centering
    \includegraphics[width=1.0\linewidth]{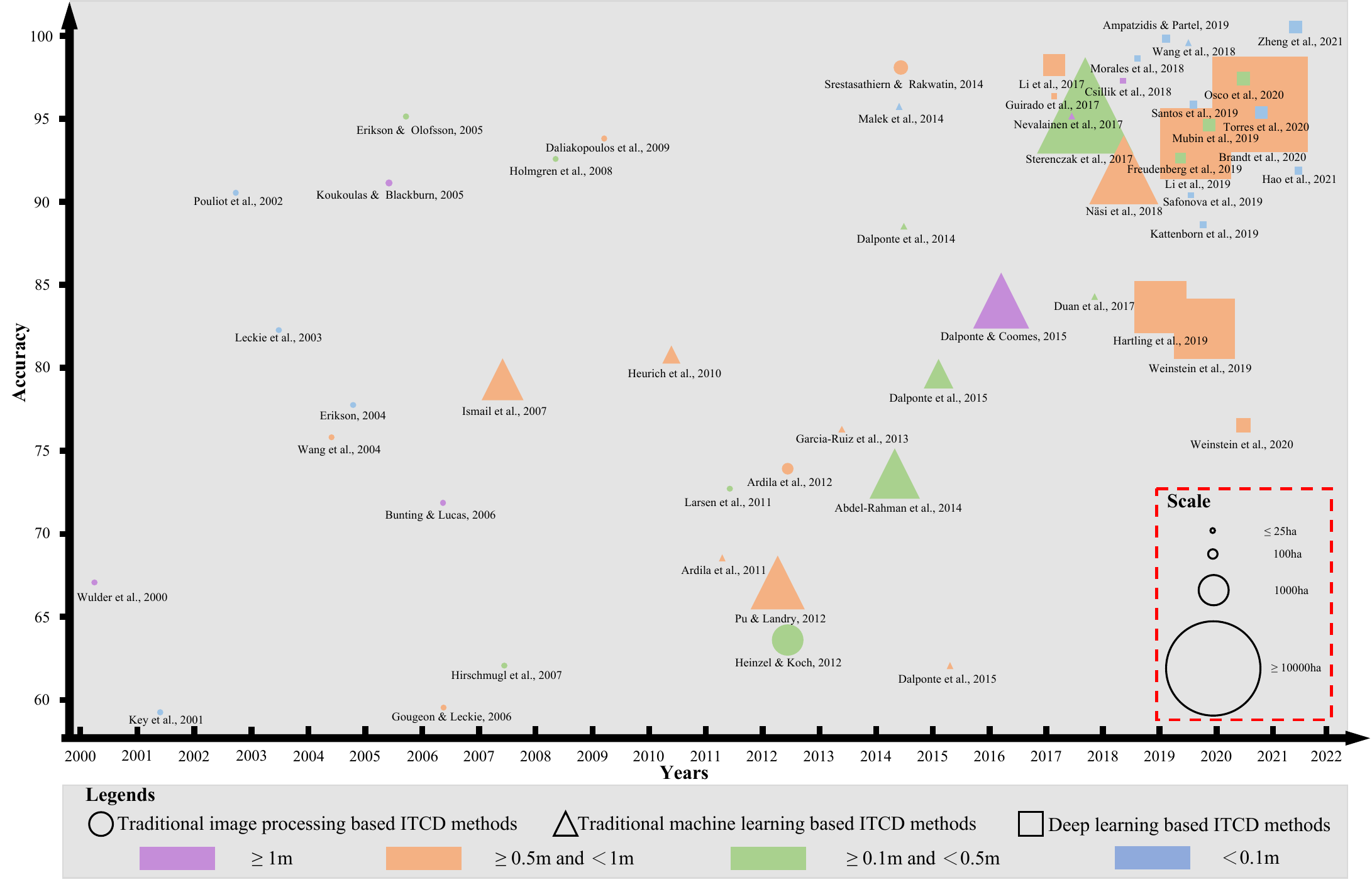}
    \caption{The overall trend of ITCD development from some typical examples since 2000. Different shapes represent different ITCD methods, and different colors represent the different spatial resolution of optical images. The larger the size is, the larger the study area is.}
    \label{fig:development}
\end{figure}

\subsection{Data collection}

Comprehensive research of existing ITCD publications is conducted in Web of Science to collect high-quality research related to ITCD, and Scopus and Google Scholar are two supplementary sources. It is noteworthy that only peer-reviewed journal publications are considered in our ITCD-related paper database. %Firstly, the database results are selected and screened through the title and abstract. Secondly, the filtering articles are further evaluated whether they sufficiently include the case study site, study area, tree species, sensor type, spatial resolution, ITCD method and evaluation, etc. 
On the foundation of screening titles and abstracts of 325 papers, an ITCD-related paper database is created with 11 fields, which served as the basis for our meta-analysis. Table \ref{tab:database} presents the examined fields in this review.  Noticeably, we display collected papers since 2000 in Fig. \ref{fig:total} and 9 collected papers published before 2000.
%\citep{gougeon1995comparison,gougeon1995crown,dralle1997automatic,brandtberg1997towards,larsen1998optimizing,uuttera1998determination,brandtberg1998automated,lowell1998evaluation,brandtberg1999automatic}. 

We categorize ITCD methods into three classes, i.e., traditional image processing-based ITCD method, traditional machine learning-based ITCD method and deep learning-based ITCD method. Fig. \ref{fig:total} also displays the number of publications per ITCD method from 2000 to 2021. Before 2005, ITCD-related publications adopted traditional image processing-based ITCD methods. Following that, with the development of machine learning techniques, traditional machine learning-based ITCD methods have gradually emerged, while traditional image processing-based ITCD methods were still in a leading position.  Since 2017, with the widespread of deep learning applications and the high performance of ITCD results, deep learning-based ITCD methods have become the most popular ITCD approach, which holds 60.9\% of publications after 2020 (78 deep learning-based ITCD publications of all 128 ITCD publications using optical related remote sensing data).%zhe li xu yao tian jia xin wen xian

\subsection{Overall trend of ITCD development}

Fig. \ref{fig:development} displays the representative ITCD-related research from 2000 to 2022. The circles, triangles and rectangles denote traditional image processing-based ITCD methods, traditional machine learning methods and deep learning-based ITCD methods, respectively. Different colors represent different spatial resolutions of remote sensing data. In addition, the larger the size is, the larger the study area is. From Fig. \ref{fig:development}, we can observe some tendencies in ITCD field:

\begin{enumerate}
    \item We can find that the left of Fig. \ref{fig:development} is sparse, while the right of holds a dense distribution, especially in the top-right corner. Since 2017, the number of ITCD-related research have been exponentially increased with high-accuracy performance.
    \item In terms of the methodology and their accuracy, ITCD studies proposed earlier have a relatively lower accuracy. Some of them have high precision ($> 90\%$ ) but their study area is really small ($< 25ha$). In addition, traditional image processing-based ITCD methods are in the majority before 2010. After that traditional machine learning-based ITCD methods gradually developed while recently, deep learning-based ITCD methods continuously emerged along with high-accuracy results. 
    \item ITCD publications focus on small study areas before 2010, most of which are smaller than $100ha$. In pace with the easier production of very-high-spatial resolution remote sensing imagery, the higher performance of computing resources, and the more robust artificial Intelligence algorithms,  more and more large-scale ITCD research has been proposed in recent years.
    \item Although adopting very-high-resolution remote sensing data ($<0.1m$, blue objects in Fig. \ref{fig:development}) promotes high-accuracy ITCD performance ($>95\%$), their study areas are quite small because of high storage costs. Most large-scale research utilizes remote sensing data with the spatial resolution of $0.5 \sim 1m$, both considering higher accuracy and lower data storage.

\end{enumerate}

\subsection{Quantitative analysis}

Along with the collection of ITCD-related publications and completion of the paper database, quantitative data are generated and presented through figures in the following subsections, including tree species, study sites and area, the types of optical remote sensing data, etc.

% \begin{figure}[t]
%     \centering
%     \includegraphics[width=1.0\linewidth]{Figures/journal_v1.pdf}
%     \vspace{-2em}
%     \caption{The number of relevant publications per literature source. We only display those the number of relevant publications is equal or larger than 3.}
%     \label{fig:journal}
% \end{figure}

\begin{figure}[t]
    \centering
    \includegraphics[width=1.0\linewidth]{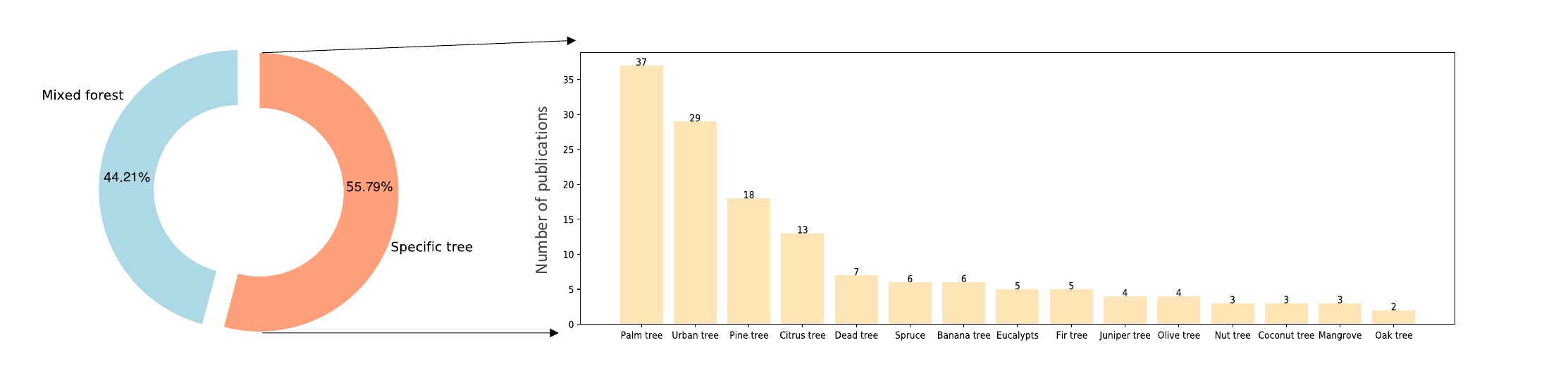}
    \caption{The statistics of tree species in ITCD-related publications. In specific trees, we only display the species that have been studied at least two times in ITCD-related papers.}
    \label{fig:species}
\end{figure}

\begin{figure}[t]
    \centering
    \includegraphics[width=1.0\linewidth]{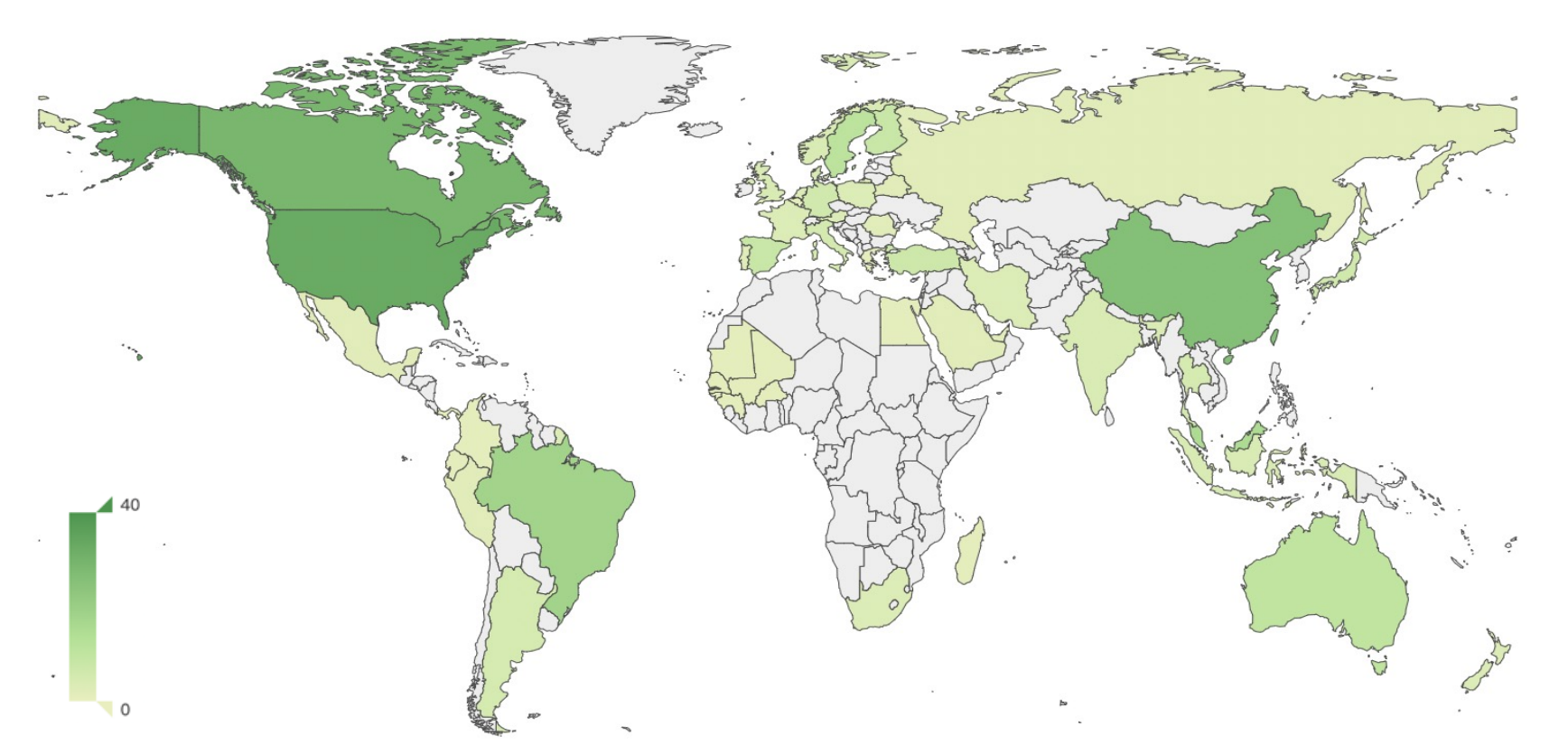}
    \caption{The number of study sites around the world according to our database.} 
    \label{fig:study_institution}
\end{figure}

% \subsubsection{Literature source}

% As displayed in Fig. \ref{fig:journal}, most of ITCD related publications (76.5\%) are derived from 20 peer-reviewed journals, while the rest are obtained from various journals. The top journal with the most number of ITCD related articles is REMOTE SENS-BASEL (18.4\%), which has assigned some special issues on the ITCD domain, such as \href{https://www.mdpi.com/journal/remotesensing/special_issues/applications_individual_tree_detection}{Applications of Individual Tree Detection (ITD)}, \href{https://www.mdpi.com/journal/remotesensing/special_issues/Individual_Tree_Detection_Characterisation_UAV}{Individual Tree Detection and Characterisation from UAV Data} and \href{https://www.mdpi.com/journal/remotesensing/special_issues/Mapping_Tree_Species}{Mapping Tree Species Diversity}, etc. Other journals that publish over 10 ITCD related papers include INT J REMOTE SENS (27 papers), REMOTE SENS ENVIRON (19 papers) and ISPRS J PHOTOGRAMM (19 papers). We can find that REMOTE SENS-BASEL has published the most number of papers that proposed deep learning based ITCD methods (25 papers) and machine learning based ITCD methods (8 papers), and INT J REMOTE SENS has published the most number of papers that proposed traditional image processing based ITCD methods (22 papers). %According to the trend of the above journals, REMOTE SENS-BASEL, ECOL INFORM and SENSORs-BASEL prefer publishing ITCD related papers using deep learning methods. At the meantime, REMOTE SENS ENVIRON, INT J REMOTE SENS  and PHOTOGRAMM ENG REM S prefer publishing ITCD related articles adopting traditional image processing approaches. 

\subsubsection{Tree species} \label{sec:species}

Fig. \ref{fig:species} displays the statistics of tree species in ITCD-related publications. In the specific tree, we only show the species that have been studied at least two times in ITCD-related papers. According to existing ITCD publications, 44.21\% of them take mixed forest as the study objective and the rest only take specific tree species as their study objective. Traditional image processing-based ITCD methods have been adopted most times when the study object is mixed forest (63.72\%). The palm tree is the most popular study species among other single tree species (37 times). The most probable reasons include the benefit of positive economics and the impact of a negative environment as the increasing expansion of oil palm plantation areas in tropical developing countries \citep{santika2021impact}. The urban tree is another popular study objective in the ITCD domain (29 times). Other popular study species include pine tree (16 times) and citrus tree (13 times).

% \subsubsection{Research institution}

% In the left of Fig. \ref{fig:study_institution}, it can be observable that according to the number of countries, research institutions are mainly located in North America, East Asia and Europe, while according to the number of papers, countries ranked top 3 are China (36 papers), USA (30 papers) and Canada (27 papers). Generally, according to Fig. \ref{fig:study_institution}, research institutions focus on study sites located in their own countries, such as Australia, Turkey and Canada, etc.,  where over 95\% of study sites in their publications are located in their own countries. On the other hand, only 56\% and 69\% of ITCD related publications consider study sites in their own countries for Italy and China, respectively.   

\subsubsection{Study sites}

Fig. \ref{fig:study_institution} displays the spatial distribution of study sites according to our database. Study sites of countries where the number is over 20 times are the USA (38 times), China (34 times) and Canada (30 times). Similar to the spatial distribution of research institutions, most of them are principally located in North America, East Asia, and North Europe. Meanwhile, tropical forest areas (such as Brazil) are a hot study site for ITCD research because of their substantial impact and significance on global climate change. %Also, as the rapid expansion of oil palm plantations is criticized by some environmental researches or institutes owing to reducing the tropical rain forest, threatening the survival of native species and destroying the biodiversity  \citep{meijaard2020environmental}, it is also a hot research region for Southeast Asia, such as Malaysia and Indonesia. 
For others such as Africa, although it has significant research value and a large distribution of tropical forests, the number of research times is quite low because of its complicated topography, lots of clouds, scarce fieldwork, and poor photograph conditions.

\begin{figure}[t]
    \centering
    \includegraphics[width=0.9\linewidth]{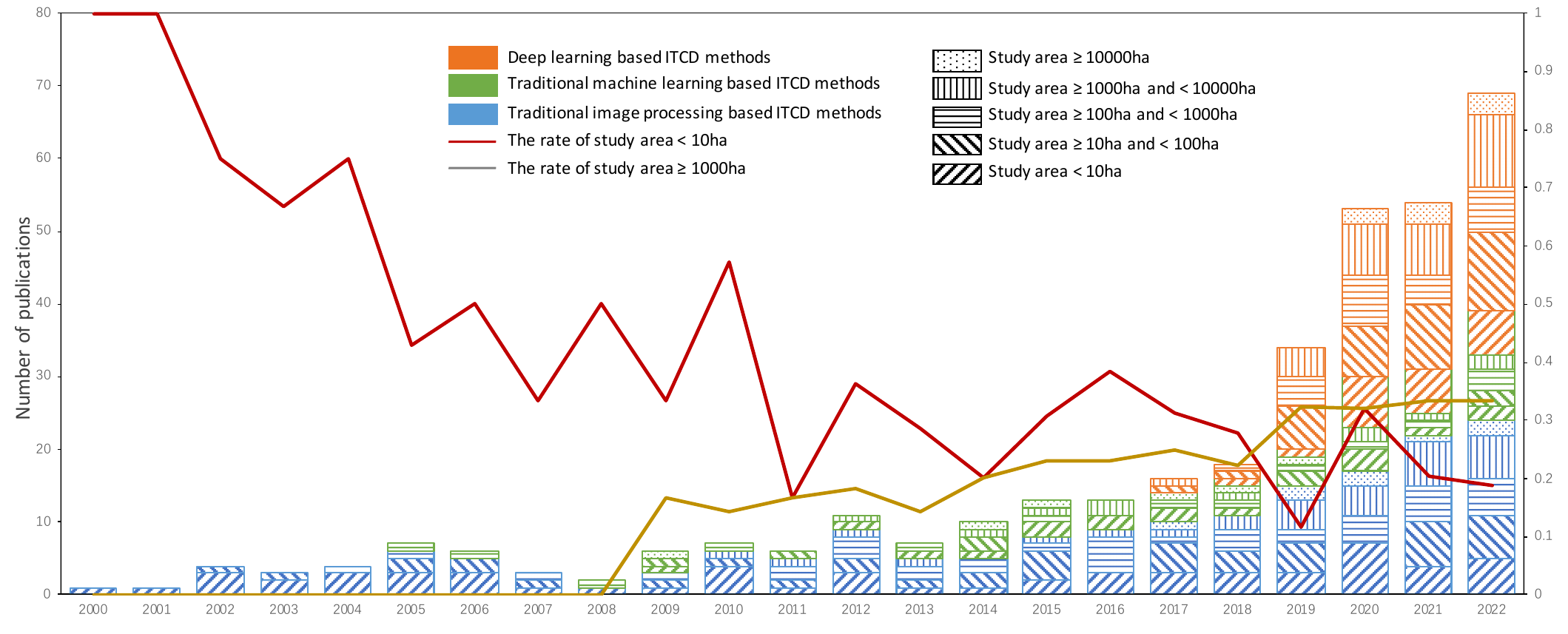}
    \caption{The statistics of study areas in ITCD-related publications. Different textures denote different study area and different colors denote different ITCD methods. The red line represents the rate of study area $\leq$ 10ha and the grey line represents the rate of study area $\geq$ 1000ha.}
    \label{fig:study_area}
\end{figure}

\begin{figure}[t]
    \centering
    \includegraphics[width=1.0\linewidth]{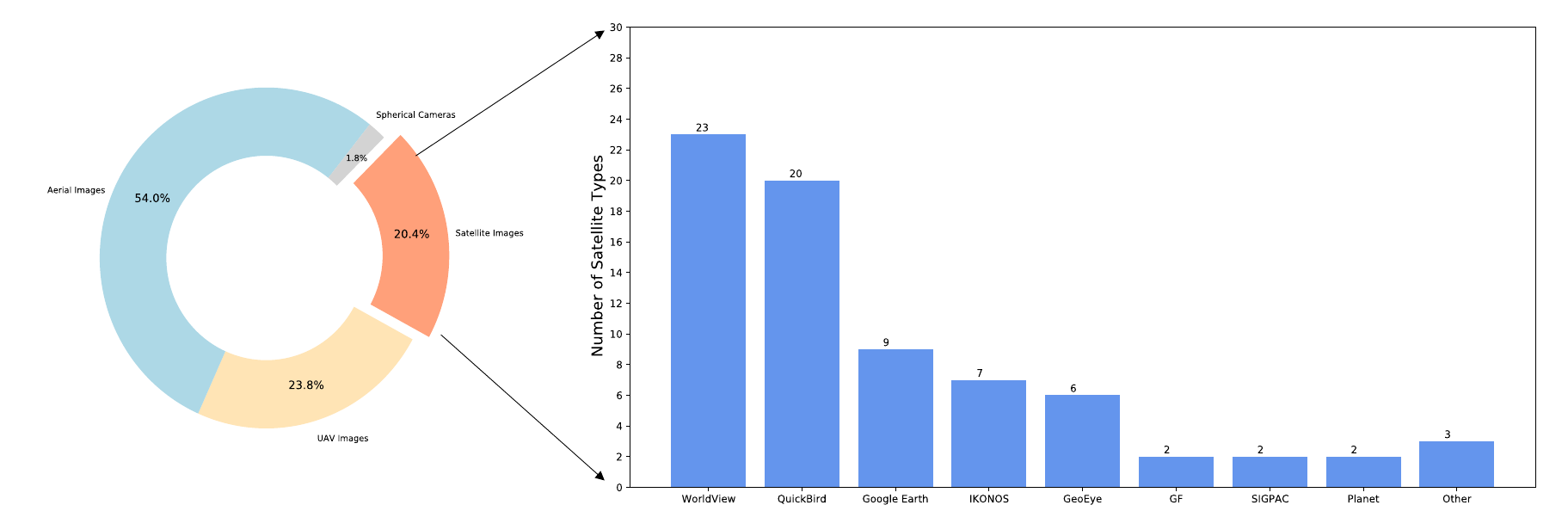}
    \caption{The number of sensor types used in ITCD-related publications, with displaying the kinds of satellite images that have been used at least 2 times in ITCD-related papers.}
    \label{fig:sensor}
\end{figure}

\subsubsection{Study area}

As Fig. \ref{fig:study_area} displays, we count the different study areas in our collected papers. Different textures denote different study area and different colors denote different ITCD methods. The red line represents the rate of study area $\leq$ 10ha and the grey line represents the rate of study area $\geq$ 1000ha. It can be seen that before 2010, the majority of the study areas were smaller than 10ha. While after 2010, the percentage of study areas larger than 1000ha are steadily increasing. 
%Specifically, 51 study areas are between 1ha and 10ha. 26.1\% of study areas are larger than 10ha yet less than 100ha, and 22.1\% of study areas locate in 100ha to 1000ha. Besides, 9.7\% of papers focus on small regions limited to 1ha and 13.7\% of papers study on large areas between 1000ha and 10000ha. 
Only 5.8\% of papers' study areas are beyond 10000ha and over 60\% of them adopt deep learning-based ITCD methods. We can observe that traditional image processing-based ITCD methods are mainly applied to study areas smaller than 100ha (on the bottom of Fig. \ref{fig:study_area}) and deep learning-based ITCD methods are more applied in larger study areas ($\geq$ 100ha) (on the left top of Fig. \ref{fig:study_area}).  We can also infer that the larger the study area is, the more deep learning-based methods are adopted, which demonstrates that deep learning-based ITCD methods generally have a stronger capacity for efficiency, generalization, and robustness.

\subsubsection{Sensor type}

Fig. \ref{fig:sensor} shows the number of sensor types used in ITCD-related publications, displaying the kinds of satellite images that have been used at least two times in ITCD-related papers.  Over half of ITCD-related publications adopt aerial images (54.0\%). %Following that, UAV images (23.8\%) and satellite images (20.4\%) are also utilized in ITCD researches. 
Recently, spherical cameras have begun to be applied in the ITCD domain, such as cameras with fisheye \citep{pearse2020detecting}, Google Street images \citep{lumnitz2021mapping}, etc. As for satellite images, it can be distinguished that WorldView and QuickBird data are adopted 23 and 20 times for ITCD applications, respectively, and present the top two places among other satellite sensor types. The total amount of the studies illustrated in Fig. \ref{fig:sensor} is larger than the number of papers examined through satellite images, indicating that data from more than one type of sensor are utilized in some publications. 

\begin{figure}[t]
    \centering
    \includegraphics[width=0.9\linewidth]{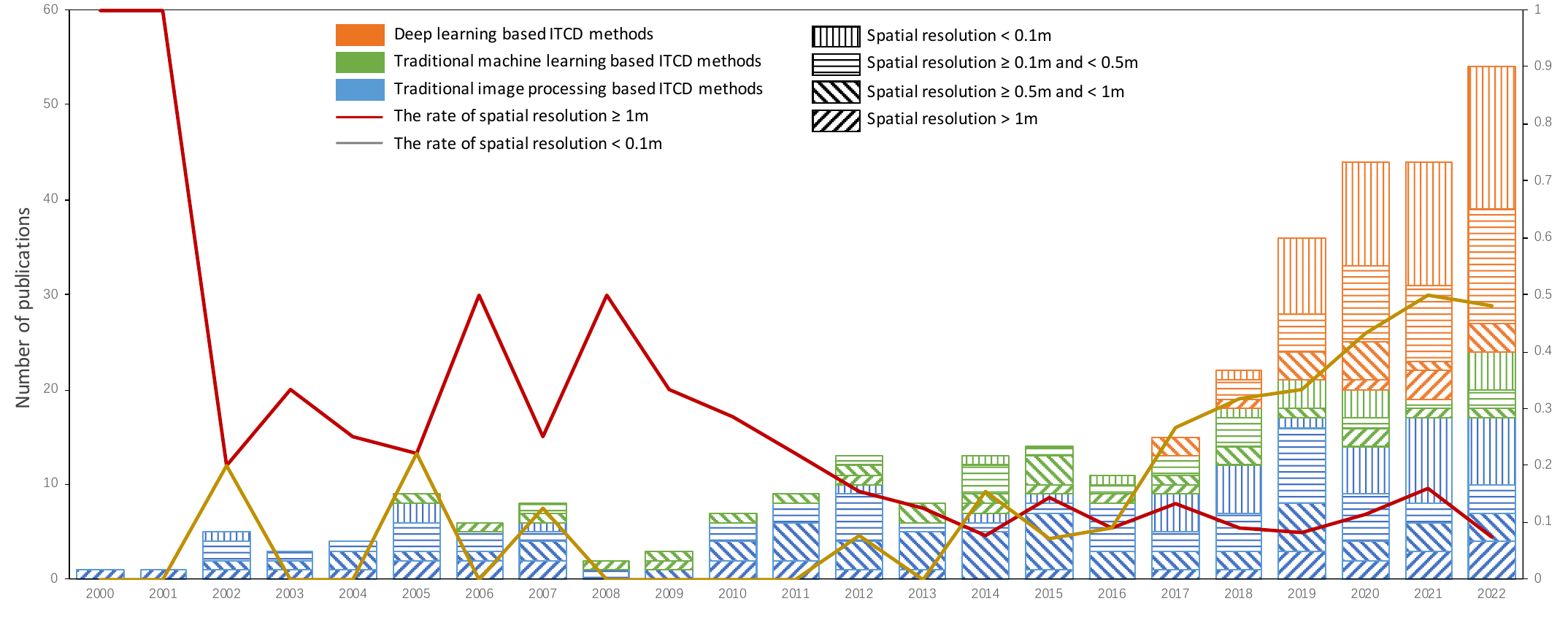}
    \caption{The statistics of spatial resolution of images used in ITCD-related publications. Different textures denote different spatial resolutions and different colors denote different ITCD methods. The red line represents the rate of spatial resolution $\geq$ 1m and the grey line represents the rate of spatial resolution $\leq$ 0.1m.}
    \label{fig:resolution}
\end{figure}

\begin{figure}[t]
    \centering
    \includegraphics[width=0.5\linewidth]{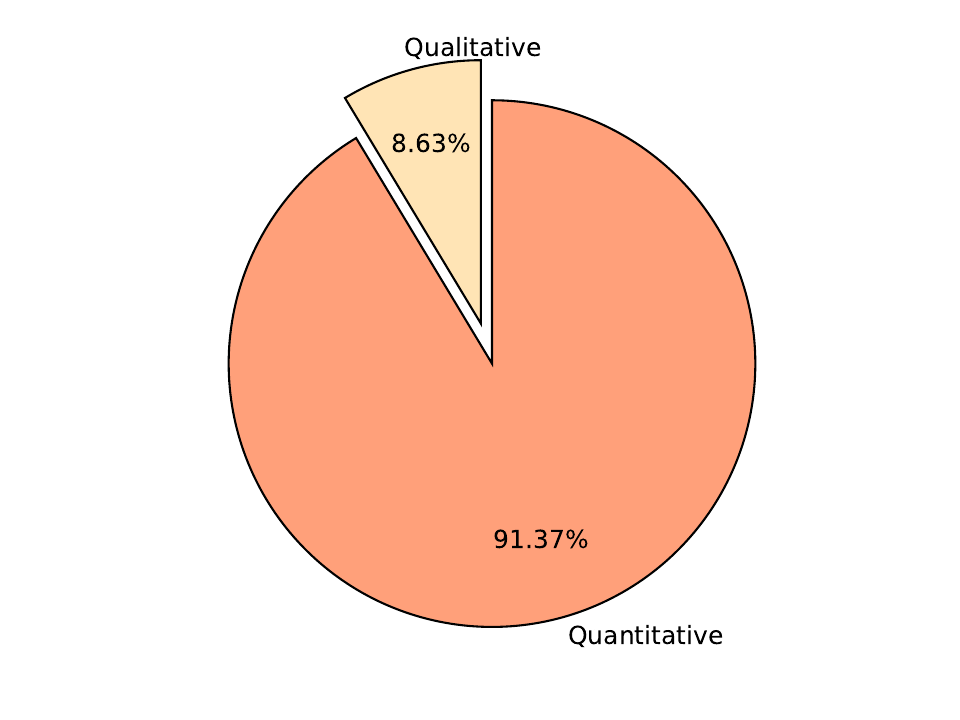}
    \vspace{-1em}
    \caption{The distribution of quantitative and qualitative ITCD evaluation methods in our collected papers. }
    \label{fig:evaluation}
\end{figure}

\subsubsection{Spatial resolution of data}

It can be seen from Fig. \ref{fig:resolution} that researchers use a very high spatial resolution of data in the ITCD domain. Different textures denote different spatial resolutions and different colors denote different ITCD methods. The red line represents the rate of spatial resolution $\geq$ 1m and the grey line represents the rate of spatial resolution $\leq$ 0.1m. We can observe that before 2010, the data with spatial resolution $\geq$ 1m was widely used in many ITCD-related papers, while the rate of spatial resolution $\leq$ 0.1m was exponentially increasing, especially after 2016. The most probable reason is that UAV images have been extensively used in forest inventory. 
As we can see, ITCD-related papers mainly focus on individual tree detection using high-resolution images. Furthermore, though traditional image processing-based ITCD methods are still the majority, with the increase of spatial resolution, more deep learning-based ITCD methods are employed, and that can be summarized as deep learning-based ITCD methods own more advantages in very-high-resolution image-based individual tree detection with stronger feature extraction and higher accuracy.

\subsubsection{Evaluation of ITCD}
\label{sec:evaluation}

%Even though a few quantitative ITCD evaluation methods have been presented in the literature, several studies adopt the trial-and-error approach to assess the results of segmentation. Thus, a more determined effort is needed to facilitate the implementation of quantitative methods. 

As can be seen in Fig. \ref{fig:evaluation}, most ITCD publications embrace quantitative evaluation, and only 8.63\% of them adopt qualitative evaluation. Those adopted qualitative evaluations mainly occur in traditional image processing-based ITCD methods and traditional machine learning-based ITCD methods. In this review, we will introduce a detailed quantitative assessment of ITCD in Sec. \ref{sec:assessment}.

\section{Methodology review}

\label{sec:method}

This section reviews the development and summary of ITCD methodology. We categorize existing ITCD methods into three classes, including traditional image processing-based ITCD method, traditional machine learning-based ITCD method and deep learning-based ITCD method. We further categorize existing deep learning-based ITCD methods into two sub-classes, including object detection-based ITCD method and semantic segmentation-based ITCD method. 

\subsection{Traditional image processing-based ITCD methods}

\begin{figure}[t]
    \centering
    \includegraphics[width=1.0\linewidth]{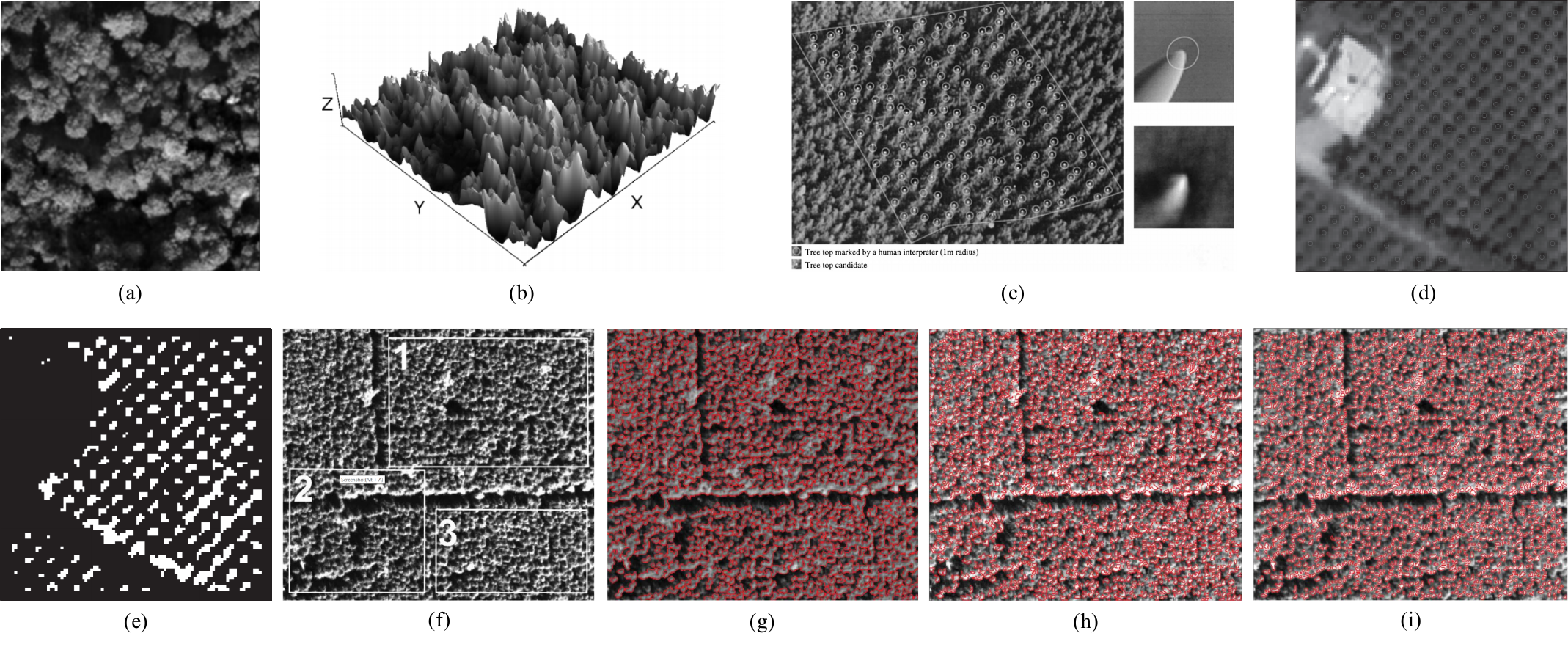}
    \caption{Some typical examples of traditional image processing-based ITCD methods. (a) original image from \citet{gomes2016detection}; (b) the local maxima appearing in the third dimension are associated with the presence of trees \citep{gomes2016detection}; (c) identification of trees through template matching from \citet{larsen1998optimizing}; (d) red band thresholding from \citet{daliakopoulos2009tree}; (e) tree detection results through image binaryzation \citep{daliakopoulos2009tree}; (f) original QuickBird image from \citet{ke2011comparison}; (g)-(i) tree delineation results using valley-following, region growing and watershed segmentation, respectively \citep{ke2011comparison}.}
    \label{fig:TIPITCD}
\end{figure}

\begin{table}[t]
    \centering
    \caption{The summary of traditional image processing based ITCD methods}
    \resizebox{\textwidth}{!}{
    \begin{tabular}{cccc}
    \hline
     Tasks &  \multicolumn{2}{c}{Methods}  &  Examples \\ \hline
     \multirow{6}*{Detection} & \multicolumn{2}{c}{Local maximum filtering}   & \citet{wulder2000local};  \citet{xu2021individual}\\ 
     & \multicolumn{2}{c}{Template matching} & \citet{niccolai2010decision}; \citet{leckie2016identifying}   \\ %\hline
     & \multicolumn{2}{c}{Image binaryzation}  &  \citet{daliakopoulos2009tree}; \citet{shafri2011semi}\\ 
     & \multirow{3}*{Others} & Scale-space filtering  & \citet{brandtberg1999automatic}; \citet{wang2010multi}\\
     &  & Object-based image analysis & \citet{bunting2006delineation}; \citet{chemura2015determination}  \\ 
     & & Marked point process & \citet{zhou2013mapping}; \citet{gomes2018individual} \\ \hline
    \multirow{3}*{Delineation} &  \multirow{2}*{Image}  & Watershed segmentation &  \citet{jing2012individual};  \citet{zheng2021mapping} \\
    & \multirow{2}*{segmentation} & Region growing & \citet{erikson2004species}; \citet{gartner2014object} \\
    & & Valley following & \citet{leckie2003stand}; \citet{leckie2005automated} \\ \hline
      %\hline
    \end{tabular}
    }
    \label{tab:tip}
\end{table}

%   ; 

Traditional image processing-based ITCD methods mainly include local maximum filtering, image binarization, template matching, object-based image analysis, image segmentation,  etc. According to previous review \citep{ke2011review} and their tasks, they can be categorized into two major types: tree crown detection and tree crown delineation (see \ref{tab:tip}). The former four methods major in tree crown detection tasks while image segmentation majors in tree crown delineation tasks. Table \ref{tab:tip} lists traditional image processing-based ITCD methods and collected examples. Fig. \ref{fig:TIPITCD} displays some typical examples of traditional image processing-based ITCD methods from existing literature.  

\textbf{Tree Detection.} Local maximum filtering premise that the presence of tree crown centers is located at the local maximum reflectance, where $\mathcal{I}(x,y)$ represents the pixel value of the image at $(x,y)$, and local value $\mathcal{L}$ is computed with $\mathcal{I}(x,y)$ in a defined window $\mathcal{A}$ to identify the local maximum. This simple and efficient method soon became the most common treetop detection approach among the traditional image processing-based ITCD methods \citep{hirschmugl2007single, van2013directional, pouliot2005approaches}. \citet{pouliot2002automated} employed a method based on local maxima and concentrated on the regeneration of coniferous forests, which resulted in the development of a seven-step procedure known as LMRDA (Local Maximum Refinement and Delineation). This procedure incorporates a "Transect" approach, involving both iterative and user-defined threshold calculations.%\citep{dralle1997automatic,wulder2000local,pitkanen2001individual,wulder2002error,wulder2004comparison,nelson2005techniques,korpela2006performance,hirschmugl2007single,lee2007detection,gebreslasie2011individual,katoh2012improving,santoro2013tree,srestasathiern2014oil,panagiotidis2017determining,ok2018combining,yi2018counting,li2019real,sun2019extraction,xiao2019individual,albuquerque2020remotely,chen2021assessment,xu2021individual,tahar2021individual}. 
%Local maximum filtering is also the foundation (the first step) of many other ITCD approaches. Some researchers developed the local maximum filtering methods to improve the performance in more complex regions or adapt to specific regions, such as local maximum refinement and delineation algorithm \citep{pouliot2002automated}, directional local filtering \citep{van2013directional}, and local maximum smoothing relation \citep{pouliot2005approaches}, etc. 
Another important branch is template matching, which recognizes trees by calculating the similarity between the templates (ground-truth trees) $T(x,y) \in \mathbf{R}^{M \times N}$ and the image patches $\mathcal{I}(x,y)$ that probably contain tree crowns \citep{solano2019methodology}. At each position $(i,j)$ in the image patch, the similarity score can be calculated normally by Sum of Squared Differences $SSD = \sum_{x=0}^{M-1} \sum_{y=0}^{N-1}[T(x,y)-\mathcal{I}(i+x,j+y)]^2$ or other metrics \citep{larsen1998optimizing, niccolai2010decision, gomes2018individual}. By doing so, the position where the similarity score is highest corresponds to the location where the template best matches the image patch, which can be recognized as the target tree \citep{niccolai2010decision, hung2012multi, gomes2018individual, huo2020individual}. 
%\citep{larsen1998optimizing,olofsson2006tree,niccolai2010integration,gulbe2013automatic,leckie2016recognition,leckie2016identifying,dos2017estimating,vahidi2018mapping,solano2019methodology,norzaki2019comparative,huo2020individual}. Template matching has been developed into some new approaches, such as multi-dimensional template matching \citep{niccolai2010decision}, multi-class template matching \citep{hung2012multi} and marked point process-based template matching \citep{gomes2018individual}, etc. 
\citet{hung2012multi} proposed a combination of template matching using features at different vision levels to overcome the detection challenges in open areas. \citet{gomes2018individual} proposed a marked point process-based template matching with incorporating a tree crown radius variable to be adaptive to the tree size. Image binarization mainly classifies the image patches into two types, i.e., tree crown and background, through threshold or filtering, which is also named image thresholding \citep{koc2018automatic}. Object-based image analysis is also widely because of the improved performance in complex scenarios  \citep{nik2021unlocking, suarez2005use}. During the initial segmentation and low-level feature extraction, object-based image analysis detects trees through the segmented images.%\citep{suarez2005use,bunting2006delineation,wu2010tree,johansen2014mapping,guerra2017use,selim2019semi}. 
%\citep{pitkanen2001individual,daliakopoulos2009tree,shafri2011semi,zhang2012individual,aliero2014usefulness,aval2018detection,khan2018remote,koc2018automatic,marques2019uav,borlaf2019methods,waleed2020automated,aluri2021framework}. 

\textbf{Tree Delineation.} Image segmentation methods mainly refer to morphological approaches, which are basically comprised of two major operations: dilation $\mathcal{D}$ and Erosion $\mathcal{E}$. Dilation is used to expand the tree regions, making them more connected and complete. Erosion can be applied to refine the tree boundaries by removing small isolated pixels or noise. Given an input image $\mathcal{I}$ , structured element $\mathcal{B}$, and coordinate $(x,y)$, dilation and erosion can be formulated as:
\begin{equation}
    \mathcal{D}(x,y) = max_{(s,t) \in \mathcal{B}}[\mathcal{I}(x-s,y-t)], \ \mathcal{E}(x,y) = min_{(s,t) \in \mathcal{B}}[\mathcal{I}(x+s,y+t)], \label{1}
\end{equation}
where $(s,t)$ means the coordinate of $\mathcal{B}$.
To delineate tree crowns, works developed morphological-based methods including watershed segmentation \citep{pouliot2005development, wang2004individual}, %\citep{wang2004individual,pouliot2005development,jing2012individual,ardila2012context,jiang2012individual,karlson2014tree,lin2015use,santoso2016simple,yang2016delineating,huang2018individual,biswas2020delineation,qiu2020new,tong2021delineation,zheng2021mapping}, 
region growing \citep{erikson2004species,gu2021individual}, %\citep{culvenor2002tida,erikson2003segmentation,erikson2004species,whiteside2011extraction,park2011automated,park2014unconstrained,yin2015object,heenkenda2015mangrove,puliti2018tree,nordin2019individual,dalponte2019individual,miraki2021individual}, 
and valley following \citep{leckie2003combined, gougeon1995comparison}, etc. \citet{wang2004individual} proposed a two-stage unified framework with marker controlled watershed segmentation to conduct tree delineation. The local maximum filtering was employed to be a marker to guide the watershed segmentation. \citet{gu2021individual} combined over-segments as the growing units to overcome the noise effects while considering the spatial and contextual information. 

Note that some research proves that combining methods, even with machine learning or deep learning-based methods may perform better ITCD results \citep{heenkenda2015mangrove}. For instance, \citet{weinstein2019individual} first uses local maximum filtering to create a big set of noisy training samples for training deep learning models, which are then finetuned by hand-crafted labels. \citet{pu2023new} designs a new combination method, involving watershed segmentation to first segment individual trees, and a k-Nearest Neighbor classifier to refine the final outputs.
\citet{pitkanen2001individual} combines locally adaptive binarization and local maximum filtering methods to achieve individual tree detection in digital aerial images, with 70-95\% of the trees detected in sparse regions. \citet{panagiotidis2017determining} combines the local maximum filtering and inverse watershed segmentation to estimate the crown diameters, achieving an acceptable accuracy for detecting tree crown diameters. \citet{weinstein2019individual} first uses local maximum filtering to create a big set of noisy training samples for training deep learning models, which are then finetuned by hand-crafted labels.
%Furthermore, some literature points out that integrating multi-source data can improve the performance of ITCD, for example, digital surface model \citep{chen2021assessment}, other geographic information system data \citep{aval2018detection} and so on.

These traditional image processing-based methods all are dependent on manual threshold selection and have difficulties with noise images, whose poor generalization and loss of fine-grained information limit the applicability. Despite these drawbacks, these methods are still significant and popular for their simplicity and high efficiency in saving time and labor consumption, reproducibility, etc. The combination of these traditional image processing-based methods with deep learning methods also brings new insights into providing fast end-to-end and convergence speed.

\subsection{Traditional machine learning-based ITCD methods}

\begin{table}[t]
    \centering
    \caption{The summary of feature extraction and adopted classifiers in the traditional machine learning-based ITCD methods}
    %\resizebox{\textwidth}{!}{
    \begin{tabular}{ccc}
    \hline
       Items  & Methods & Examples \\ \hline
        & \multirow{2}*{Non-handcrafted features}  & \citet{tooke2009extracting}; \citet{dalponte2014tree}; \\
       Feature &    & \citet{nevalainen2017individual}; \citet{johansen2020mapping} \\
       extraction & \multirow{2}*{Handcrafted features}  & \citet{ouma2008urban}; \citet{pu2012comparative}; \\
       & & \citet{malek2014efficient}; \citet{dalponte2015semi} \\
       \hline
       
        & Decision tree & \citet{tooke2009extracting}; \citet{tochon2015use}; \\
       & Gaussian maximum likelihood &  \citet{bai2005quantifying}\\
       & Linear discriminant analysis & \citet{pu2012comparative} \\
       %& Canonical variate analysis & \citet{ismail2007forest} \\
       \multirow{2}*{Adopted} & Support vector machine &  \citet{wang2019automatic}; \citet{windrim2020tree}\\
        \multirow{2}*{classifiers} & Extreme learning machine & \citet{malek2014efficient} \\
         & Random forest &  \citet{roth2019automated}; \citet{johansen2020mapping} \\

       & Multi-layer perceptron & \citet{nevalainen2017individual} \\
       & K-means & \citet{recio2013automated}; \citet{dalponte2015delineation}; \\
       & K-near neighborhood & \citet{heurich2010object}; \citet{mollaei2018detection} \\
       & Logistic regression & \citet{wu2018mapping} \\
       %& Bayes classifier & \citet{ozcan2020probabilistic} \\
       \hline

    \end{tabular}
    %}
    \label{tab:tml}
\end{table}

%; \citet{wang2019automatic}
%; \citet{chen2021kdt}        & & \citet{al2018image} \\\citet{ouma2008urban}; 
%\citet{dalponte2015semi}; \citet{secord2007tree}; \citet{jiang2012individual}; \citet{garcia2013comparison};  \\   & & \citet{dalponte2014tree}; \citet{lopez2016early}; \\   & &  \citet{nasi2018remote};
%\citet{dash2019early}; \citet{johansen2020mapping};\\\citet{abdel2014detecting}; \citet{duan2017novel};   \citet{white2018uas}; \\ & &        & &  \citet{tagle2020identifying}; \citet{wallace2021linking}; \citet{yu2021machine}\\    & &  \citet{apostol2020species}; \citet{man2020automatic}; \citet{selvaraj2020detection};\\
%; \citet{kestur2018tree} \citet{ouma2006optimization}; \citet{kandare2017individual}       & &  \\
%\citet{jiang2012individual}; \citet{lopez2016early};  ; \citet{windrim2020tree}

The revolution in machine learning facilitates the development of ITCD by offering powerful, adaptable, accurate solutions. Generally speaking, for both tree detection and tree delineation, there are four steps in traditional machine learning-based ITCD methods: (1) image pre-processing; (2) feature extraction; (3) classifier training; and (4) model prediction. Here we focus more on the nature of ITCD, which is the progress in feature extraction and classifier training. Table \ref{tab:tml} lists traditional machine learning-based ITCD methods and collected examples. Because for tree detection and tree delineation, feature extraction and classifier training are both necessary and employed methods are similar, this section does not separate detection and delineation. %Image pre-processing includes  atmospheric correction \citep{dalponte2014tree}, spectral normalization \citep{dalponte2015semi}, spectral reduction \citep{tochon2015use}, and cloud masking \citep{roth2019automated}, etc. 

\textbf{Feature Extraction.} There is a variety of feature extraction methods, which can be simply classified into two types, i.e., non-handcrafted features and handcrafted features. Non-handcrafted features mainly utilize obvious inner features of images themselves, such as spectral information, vegetation index  \citep{ouma2008urban}, texture characteristics (e.g., Gray-Level Co-occurrence Matrix, GLCM) \citep{pu2012comparative}, structure characteristics \citep{lopez2016early}, etc. Some studies also take spectral reflectance \citep{tooke2009extracting}, canopy height model \citep{wu2018mapping} and point cloud data \citep{kaminska2018species} into consideration. \citet{ouma2008urban} first selects suitable bands for urban trees from QuickBird images and then calculates a normalized difference vegetation index using selected bands to extract related features. \citet{kaminska2018species} extracts point-cloud features by deriving intensity and structural variables and spectral information from aerial images for dead tree detection. On the other hand, handcrafted features are specific image representations that are crafted by domain knowledge and prior understanding of the data. These features are created by specific methods (e.g., Principal Component Transform (PCT), Scale-Invariant Feature Transform (SIFT), Histogram of Oriented Gradient (HOG), etc.) to capture relevant information that is deemed important for ITCD tasks. PCT begins with centering the data $\mathcal{D}$ by subtracting the mean value $\mu$. Then the covariance matrix is calculated by:
\begin{equation}
    Cov(X_i,X_j) = \frac{1}{n-1}\sum_{k-1}^n(X_{k,i} - \hat{X}_i)(X_{k,j} - \hat{X}_j), \label{2}
\end{equation}
where $X_{\{i,j\}}$ are the $i$-th and $j$-th features, and  $\hat{X}_{\{i,j\}}$
are the mean of  $i$-th and $j$-th features. After that eigenvalue decomposition is conducted to find eigenvectors and eigenvalues, which are the crafted features.  \citet{ouma2008urban} utilized PCT to extract handcrafted features for urban tree detection. SIFT extracts features by convolving the image with Gaussian kernels $\mathcal{G}$ of different scales.  \citet{malek2014efficient} extracts a set of key points by SIFT before classifier training for palm trees. Given an input image $\mathcal{I}$, HOG calculates gradient orientations through vertical and horizontal directions 
\begin{equation}
    \Theta(x,y) = arctan2(\mathcal{I}_x(x,y), \mathcal{I}_y(x,y)), \label{3} 
\end{equation}
and then constructs histograms $\mathcal{H}(x,y)$ of these orientations. \citet{wang2019automatic} uses HOG to extract effective features for automatically detecting individual trees. 

The interpretability of these features makes them useful for understanding and reasoning about the content as they are explicitly designed to capture certain visual attributes like tree shapes and edges. Compared with the non-handcrafted features, handcrafted features present more data-driven characteristics. Still, due to the requirements of manual design and expert understanding, these features are lack of scalability and transferring ability to new scenarios. In a nutshell, a full understanding of the characteristics of the scenarios and the specific demands of specific ITCD tasks is essential to harness the full potential of these features, and is beneficial for later classifier training. 

\textbf{Classifier training.} Classifier training is the most important part in traditional machine learning ITCD methods. Potential classifiers contain Decision Trees (DT), %\citep{ouma2008urban,tooke2009extracting,tochon2015use,dalponte2016tree,al2018image,chen2021kdt}, 
Gaussian maximum likelihood, linear discriminant analysis, %canonical variate analysis \citep{ismail2007forest}, 
Support Vector Machine (SVM), Extreme Learning Machine (ELM), %\citep{secord2007tree,jiang2012individual,garcia2013comparison,dalponte2014tree,dalponte2015semi,lopez2016early,zarea2015novel,nasi2018remote,wang2019automatic,windrim2020tree}, extreme learning machine \citep{malek2014efficient}, 
Random Forest (RF),  %\citep{poznanovic2014accuracy,abdel2014detecting,duan2017novel,white2018uas,kaminska2018species,roth2019automated,dash2019early,johansen2020mapping,apostol2020species,selvaraj2020detection,tagle2020identifying,wallace2021linking,wickramarathna2021automated,man2020automatic,yu2021machine}, 
Multi-Layer Perceptron (MLP), K-means,  %\citep{ouma2006optimization,recio2013automated,dalponte2015delineation,kandare2017individual,kestur2018tree}, 
k-Nearest Neighbors (k-NN) and logistic regression,  etc. For example, DT provides an interpretable framework for decision-making by recursively splitting data based on feature values $\mathcal{X}$. Each internal node $\mathcal{N_i}$ represents a feature and a decision threshold, while each leaf node $\mathcal{N_l}$ corresponds to a class label or a prediction. \citet{ouma2008urban} uses DT after getting the NDVI features to decide whether a pixel belongs to urban trees or not, and the performance is promising compared with the parametric maximum-likelihood classifier.  To overcome the overfitting issues associated with individual DT, RF builds multiple decision trees $\hat{\mathcal{Y}}$ during training and combines their predictions $f_i(\mathcal{X})$ to enhance the randomness and diversity. Each tree is constructed using a bootstrapped subset of the training data, and at each node, a random subset of features is considered for splitting. \citet{abdel2014detecting} proposes a RF-based classifier to identify the location and class of individual trees. As for MLP, it is a fundamental architecture of neural network and consists of multiple layers of interconnected neurons. MLP can be defined as:
\begin{equation}
    o_j = \sigma(\sum_{i=1}^n\omega_{ij}x_i+b_j), \label{4}
\end{equation}
where $o_j$ is the output of neuron $j$, and $\sigma$ is the activation function. $n$ is the number of neurons, and $\omega_{ij}$ denotes the weights, while $x_i$ and $b_j$ denote the input and the bias. Introducing non-linearity and raw representation learning, MLP can outperform other classifiers when the quality and quantity of data are guaranteed.  \citet{nevalainen2017individual} compares ITCD performance of 5 different classifiers including k-NN, Bayes classifier, DT, MLP and RF. They build a high-resolution dataset based on hyperspectral and point cloud data and extract about 350 features. Experimental results indicate that MLP achieves the best accuracy with 95.4\%, following k-NN, RF, DT and Bayes classifiers.

In summary, the performance of traditional machine learning-based methods in ITCD relies on efficient feature extraction and powerful classifier training. Compared with traditional image processing-based methods, the ability to automatically learn and extract relevant features from raw data involving specific expert understanding of traditional machine learning-based methods makes them more adaptable to more different scenarios for ITCD. However, one important thing to note is that a set of high-quality and quantity input data is a sufficient condition for the promising performance of traditional machine learning-based methods. For example, if the study area is a small region with simple tree targets and landscape invariance and the images are full of noise, traditional image processing-based methods may have better performance. Therefore, it is dependent on the specific scenarios and data conditions when choosing or comparing  traditional image processing-based and traditional machine learning-based methods.

\subsection{Deep learning-based ITCD methods}

\begin{figure}[t]
    \centering
    \includegraphics[width=0.8\linewidth]{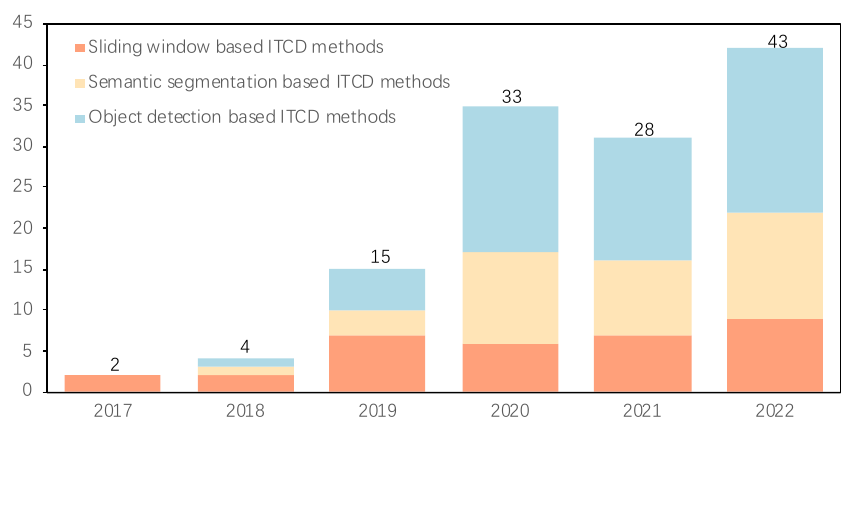}
    \vspace{-5em}
    \caption{The number of deep learning based ITCD methods related publications since 2017.}
    \label{fig:dl-method}
\end{figure}

As successful cases emerging in various applications \citep{lecun2015deep}, nowadays many ITCD methods adopt Convolutional Neural Networks (CNN), achieving high-accuracy and real-time ITCD results in complex and large-scale regions. Here we also review deep learning-based ITCD methods by the extended taxonomy: object detection-based methods for tree detection, and semantic segmentation-based methods for tree delineation. Table \ref{tab:tdl} lists deep learning-based ITCD methods and collected examples. Fig. \ref{fig:dl-method} displays the number of deep learning-based ITCD methods-related publications since 2017.

\begin{table}[t]
    \centering
    \caption{The summary of the deep learning-based ITCD methods}
    \resizebox{\textwidth}{!}{
    \begin{tabular}{ccc}
    \hline
       Methods  & Networks & Examples \\ \hline
      % \multirow{7}*{CNN classification} & LeNet \citep{lecun1998gradient}  & \citet{li2017deep};  \citet{mubin2019young};  \\

      %   &  AlexNet \citep{krizhevsky2012imagenet}  & \citet{tao2020deep}; \citet{zheng2020cross}; \\

      % & VGG \citep{simonyan2014very}  & \citet{safonova2019detection}; \citet{de2020learning}; \\

      % & GoogleNet \citep{szegedy2015going} & \citet{tao2020deep} \\
      % & ResNet \citep{he2016deep}  &  \citet{sun2019characterizing};    \citet{onishi2021explainable} \\
      % & Inception \citep{szegedy2016rethinking} &  \citet{hu2020recognition} \\
      % & DenseNet \citep{huang2017densely} & \citet{hartling2019urban} \\
       %& Others & \citet{dong2018single}; \citet{dong2019progressive}; \citet{li2019large} \\
      % \hline
             \multirow{14}*{Object detection} 
       & LeNet \citep{lecun1998gradient}  & \citet{li2017deep};  \citet{mubin2019young};  \\
        & VGG \citep{simonyan2014very}  & \citet{safonova2019detection}; \citet{de2020learning}; \\
        & ResNet \citep{he2016deep}  &  \citet{sun2019characterizing};    \citet{onishi2021explainable} \\
       & Inception \citep{szegedy2016rethinking} &  \citet{hu2020recognition} \\
       & DenseNet \citep{huang2017densely} & \citet{hartling2019urban} \\
       & YOLO \citep{redmon2018yolov3}  & \citet{ampatzidis2019citrus};  \citet{itakura2020automatic} \\
       & SSD \citep{liu2016ssd} & \citet{ple2020individual} \\
       &  RetinaNet \citet{lin2017focal}  & \citet{selvaraj2020detection}; \citet{weinstein2020cross} \\
       
       & EfficientDet \citep{tan2020efficientdet} & \citet{ammar2021deep}\\
       
       & Faster R-CNN \citep{ren2015faster}   &  \citet{pearse2020detecting};  \citet{zheng2021growing} \\
       & Mask R-CNN \citep{he2017mask}  &  \citet{ocer2020tree}; \citet{lumnitz2021mapping} \\
       & MMDetection \citep{chen2019mmdetection} & \citet{zamboni2021benchmarking} \\
       & DetectNet \citep{tao2016detectnet} & \citet{pulido2020assessment} \\
       & DeepForest \citep{weinstein2020deepforest} & \citet{aubry2021multisensor}; \citet{weinstein2020deepforest} \\
       \hline
       \multirow{6}*{Semantic segmentation} & DeepLabV3+ \citep{chen2018encoder}  & \citet{ferreira2020individual};  \citet{martins2021semantic} \\
       &  U-Net \citep{ronneberger2015u} & \citet{freudenberg2019large};     \citet{brandt2020unexpectedly}; \\

       & FCN \citep{long2015fully}  & \citet{xiao2020treetop}; \citet{osco2020semantic} \\
       & SegNet \citep{badrinarayanan2017segnet} & \citet{ochoa2019framework}\\
       & FC-DenseNet \citep{jegou2017one} & \citet{lobo2020applying} \\
       & ResNet-like \citep{he2016deep} & \citet{osco2020convolutional}; \citet{yao2021tree} \\
       %& Others & \citet{miyoshi2020novel}; \citet{qin2021identifying} \\
       \hline

    \end{tabular}
    }
    \label{tab:tdl}
\end{table}

%\subsubsection{CNN classification based ITCD methods}

\begin{figure}[t]
    \centering
    \includegraphics[width=1.0\linewidth]{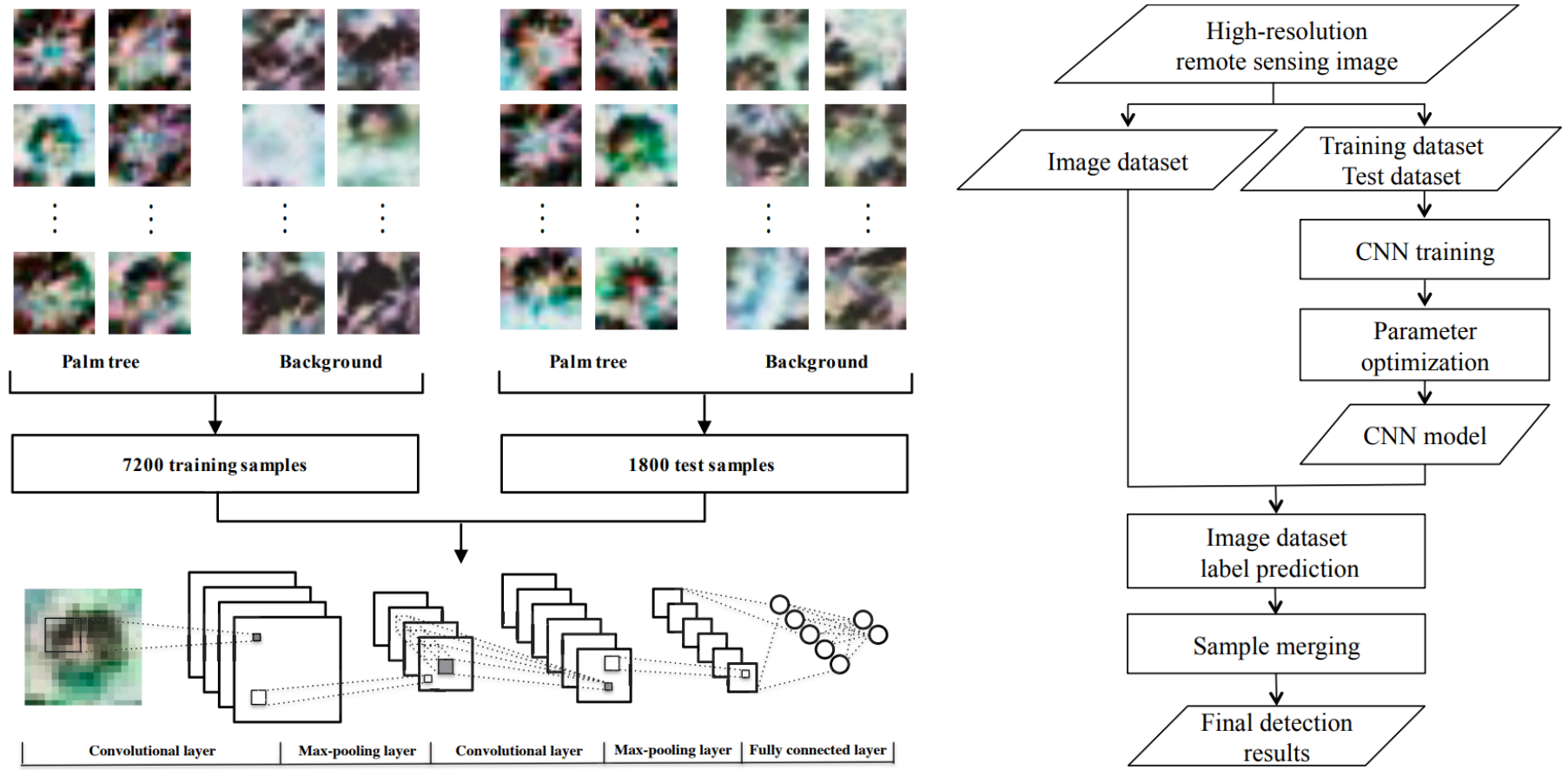}
    \caption{A typical example of CNN classification based ITCD method proposed by \citet{li2017deep}.}
    \label{fig:CNNITC}
\end{figure}
\begin{figure}[t]
    \centering
    \includegraphics[width=1.0\linewidth]{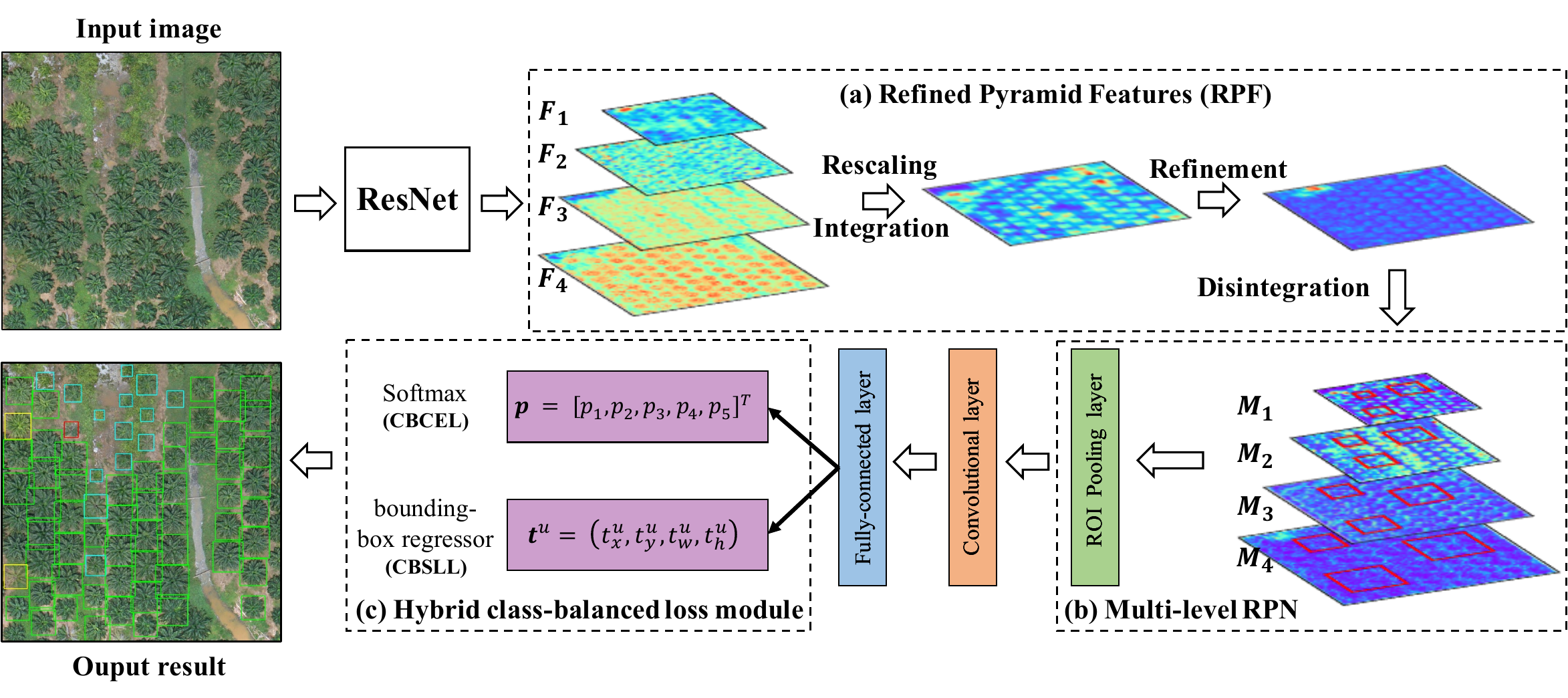}
    \caption{A typical example of object detection based ITCD method proposed by \citet{zheng2021growing}.}
    \label{fig:ODITC}
\end{figure}

\textbf{Tree Detection.} A rich line of object detection approaches have been applied to detect a variety of ground objects in remote sensing field in the past few decades, including tree detection using high-resolution remote sensing data. Object detection algorithms can be generally categorized into two classes: sliding window-based methods and end-to-end methods (i.e., two-stage object detection methods and one-stage object detection methods). Sliding-window-based object detection methods are the earliest methods in deep learning-based tree detection methods and were proposed in 2017 \citep{li2017deep} (see Fig. \ref{fig:CNNITC}). Common structures include LeNet,  %\citep{li2017deep,csillik2018identification,mubin2019young,bhattacharyya2020crown,flores2021automated,aeberli2021detection}, 
AlexNet, %\citep{wu2020domain,zheng2020cross,nguyen2021individual,zheng2021multisource}, 
VGG %\citep{safonova2019detection,de2020learning,nguyen2021individual,khan2021health},
GoogleNet, Inception, ResNet, %\citep{guirado2017deep,sun2019characterizing, onishi2021explainable,wang2021early}, 
DenseNet and other CNN structures. Many scholars have designed new CNN architectures to improve the performance of tree detection. For example, \citet{dong2019progressive} propose progressive cascaded CNN to effectively alleviate wrong detected trees and missing trees in the scene of a complex forest because of unclear canopy contour and abnormal shape. Their model attains 3.9$\sim$11\% improvement in three study areas located in China, Thailand, and America. \citet{li2019large} present a two-stage CNN architecture to detect and count oil palms in Malaysia. The first stage is to classify the land cover type and the second stage is to classify the object. Experimental results demonstrate that two-stage CNN has much fewer confusions with other land cover types (such as other vegetation and buildings) in the whole QuickBird image and achieves 21.27\% at most improvement compared to traditional one-stage CNN with respect of the F1-score. Furthermore, some researchers propose novel approaches to reduce time-consuming label interpretation work. For instance, \citet{de2020learning} proposes feature learning from image markers to largely decrease the number of training images in fully connected layers, and the accuracy has a slight improvement of 0.3\% compared to fine-tuning the VGG structure for coconut detection. \citet{zheng2020cross} introduce a new domain adaptation model based on AlexNet for cross-regional oil palm tree detection, improving the detection accuracy by 14.98\% with respect to average F1-score compared with a straightforward CNN architecture, without adding any annotations in the new study area.

On the other hand, end-to-end methods contains two-stage object detection framework and one-stage object detection framework. Two-stage object detection framework, to some extent, consists of the mechanism of the human brain, firstly giving a coarse scan of the whole image and then focusing on areas of interest. It contains several correlated stages, such as generating region proposals, CNN-based feature extraction, bounding box regression, and classification. The most common two-stage object detection-based ITCD method is Faster R-CNN. (see Fig. \ref{fig:ODITC})%\citep{neupane2019deep,xie2019detecting,zheng2019large,wu2020extracting,hu2020detection,wu2020cross,chen2020improved,pearse2020detecting,deng2020detection,culman2020individual,liu2021automatic,yarak2021oil,zheng2021growing,moura2021towards,emin2021target,xia2021automatic,zheng2021coconut}. 
In addition, Mask R-CNN, which is also a two-stage object detection method along with an instance segmentation module, has been utilized in ITCD field. %\citep{g2020tree,chiang2020deep,ocer2020tree,chadwick2020automatic,hao2021automated,lumnitz2021mapping,safonova2021olive}. 
Based on global regression/classification, a one-stage object detection framework straightly maps from image pixels to class probabilities and bounding box coordinates. One-stage object detection based ITCD methods include YOLO v2/v3,  %\citep{ampatzidis2019citrus,ampatzidis2019uav,itakura2020automatic,yuan2020fpga}, 
Single Shot Detector (SSD), RetinaNet, DetectNet and EfficientDet, etc. To this end, various wall-to-wall object detection methods have been adopted in tree detection field. \citet{weinstein2020deepforest} develop a new python package, DeepForest, to detect individual tree crowns through high-resolution remote sensing images using an object detection-based deep learning approach. This package makes the procedures of retraining and utilizing deep learning algorithms easier for a range of spatial resolutions, sensors and forests. Recent ITCD works adopt DeepForest to count and detect trees in forest area \citep{weinstein2021benchmark}. \citet{xia2019fast} conduct analysis on the efficiency and accuracy of the aforementioned algorithms applied on ITCD applications, indicating that two-stage object detection approaches (such as Mask R-CNN, Faster R-CNN) generally attain higher accuracy than one-stage object detection approaches (such as RetinaNet, YOLO v2/v3), but one-stage methods potentially accelerate the ITCD speed. \citet{santos2019assessment} compare three different object detection ITCD methods (i.e., Faster R-CNN, YOLO v3 and RetinaNet) on a UAV dataset comprising 392 RGB images over a forested urban area in midwest Brazil. Experiments indicate that RetinaNet presents the most accurate results with 92.64\% on average over all five rounds. YOLOv3 and Faster-RCNN come next, with an average precision of 85.88\% and 82.48\%. Similar to \citet{xia2019fast}, Faster-RCNN has the highest computational cost since it contains two sequential stages, while YOLO v3 and RetinaNet achieve a speed of approximately 6.3 and 2.5 times faster than Faster-RCNN, respectively, largely owing to they address object detection as a regression issue. Besides, various researchers not only are not limited to tree detection domain, but also focus on the health assessment of trees \citep{selvaraj2020detection}. For example, \citet{zheng2021growing} propose an improved Faster R-CNN model to automatically detect five fine-grained oil palm growing statuses in Indonesia, including yellowish oil palm, smallish oil palm, mismanaged oil palm, healthy oil palm and dead oil palm (see Fig. \ref{fig:ODITC}). Their work proves considerable potential not only for individual oil palm tree detection, but also monitoring of growing status using UAV images, resulting in more efficient and precise management of oil palm plantation regions. Furthermore, some publications estimate other tree parameters on the basis of ITCD results, such as the tree crown morphology \citep{g2020tree}, tree crown size \citep{ampatzidis2019citrus}, tree height \citep{hao2021automated} and tree canopy volume \citep{safonova2021olive} which are conducive to ecological assessment and management.

The first row of Table \ref{tab:tdl} summarizes the algorithms and collected examples for object detection based-methods for tree detection. In general, despite sliding-window-based methods achieving much better performance than traditional image processing-based methods and traditional machine learning-based methods in areas with crowded or overlapping trees, they have to adopt the sliding window technique to complete final results, which is a time-consuming approach due to producing a considerably large number of latent candidates ranging from a variety of sizes. Therefore, sliding-window-based methods are inflexible and inefficient to detect trees with various crown sizes since the patch size of the sub-image is required to be predefined through human prior knowledge. For example, although \citet{mubin2019young} detect mature and young oil palms, they define different sliding window sizes in advance for mature and young oil palms (31$\times$31 for mature oil palms and 26$\times$26 for young oil palms, respectively). As for end-to-end object detection-based methods such as Faster R-CNN, they are more robust and faster, greatly alleviating the performance drop caused by confusion with other vegetation or complex topography, etc. Compared to traditional machine learning-based ITCD methods and sliding-window-based methods, end-to-end object detection-based methods have a considerable improvement in accuracy and efficiency \citep{zheng2021growing}. Nowadays, end-to-end object detection-based algorithms are more and more popular and common among all ITCD methods, holding 25.0\%, 33.3\%, 50.0\% and 46.4\% of deep learning-based ITCD methods in 2018, 2019, 2020 and 2021, respectively (see detail in Fig. \ref{fig:dl-method}).%In addition, the CNN classification based methods not only need to annotate tree crown samples, but also need to annotate for other land cover classes (such as background, impervious, and other vegetation, etc.) with extra efforts. \citep{sun2019characterizing}. 

\begin{figure}[t]
    \centering
    \includegraphics[width=1.0\linewidth]{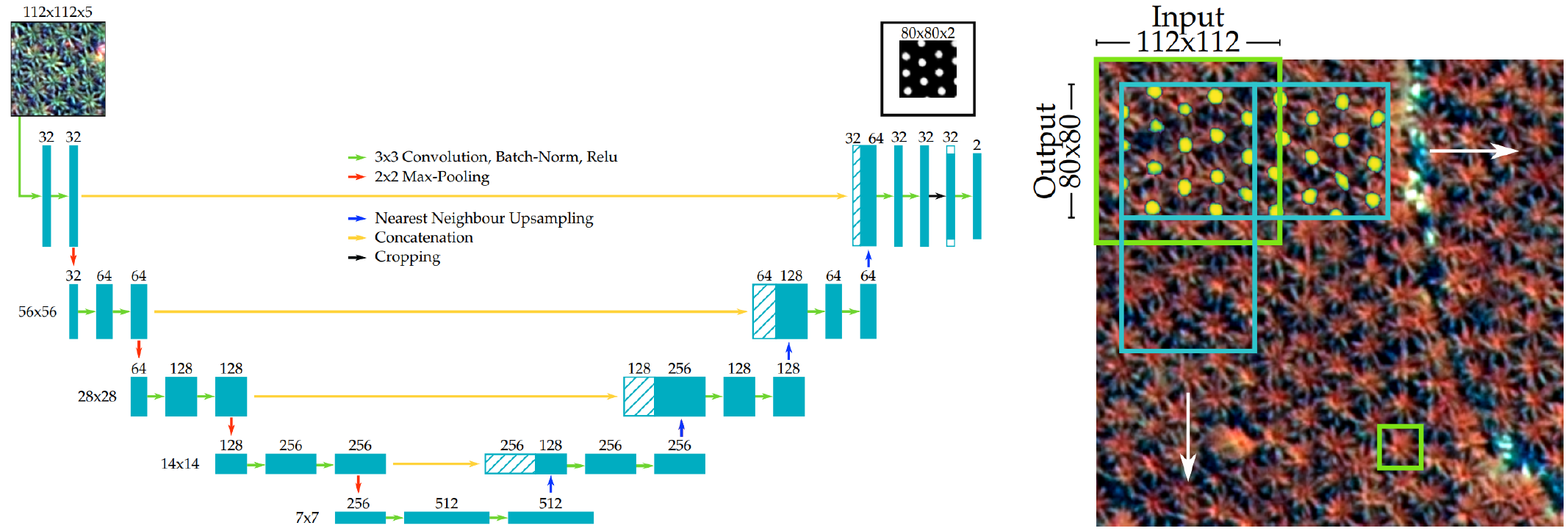}
    \caption{A typical example of semantic segmentation based ITCD method proposed by \citet{freudenberg2019large}.}
    \label{fig:SSITC}
\end{figure}

\textbf{Tree Delineation.} Without requiring the time-consuming sliding window scheme, semantic segmentation-based tree delineation method is a wall-to-wall algorithm. Dissimilar with the CNN object detection-based methods that produce one label for a patch of image, semantic segmentation methods aim at generating dense classes for each pixel in the whole image (see Fig. \ref{fig:SSITC}). Some state-of-the-art semantic segmentation architectures, such as DeepLabV3+,  %\citep{morales2018automatic,ferreira2020individual,ferreira2021accurate,martins2021semantic,cheng2021cherry}, 
U-Net, %\citep{freudenberg2019large,chen2019trees,kattenborn2019convolutional,wu2020extracting,wagner2020regional,zhang2020identifying,brandt2020unexpectedly,gibril2021deep,korznikov2021using}, 
Fully Connected Network (FCN), FC-DenseNet and SegNet, etc. have been applied to ITCD domain in recent years. Some researchers adopt semantic segmentation-like models to generate confidential map for extracting tree crowns \citep{miyoshi2020novel}. Those pixels which hold high confidence are the locations of tree crowns. Some researchers also design new algorithms to alleviate the impediment of the required volume of labelled data for training. For example, \citet{qiao2020simple} propose a weakly supervised deep learning pipeline \citep{zhou2016learning} and class activation mapping to detect individual red-attacked trees, only along with image-level-labeled remote sensing data. \citet{xiao2020treetop} presents a workflow to train FCN with automatically utilizing produced pseudo labels through unsupervised treetop detectors, which effectively saves manual labeling efforts while keeping a comparable performance. Some papers propose a modified semantic segmentation model for delineation applications. For instance, \citet{zhang2020identifying} integrates a set of residual U-Nets and a sequence of automatically derived input scales to introduce a new scale sequence residual U-Net-based deep learning algorithm, which is able to complete self-adaption to variation in different kinds of trees, consistently attaining the highest detection accuracy  (91.67\% on average)  compared with other four state-of-the-art ITCD-related approaches. \citet{qin2021identifying} presents a novel architecture, spatial-context-attention model, to recognize pine nematode disease based on UAV multi-spectral remote sensing imagery, outperforming other semantic segmentation approaches including DeepLab V3+, DenseNet, and HRNet. As for the comparison among different semantic segmentation methods, \citet{osco2020semantic} evaluates five state-of-the-art semantic segmentation-based tree delineation methods to semantic segment citrus-trees from UAV multispectral images, including DeepLabV3+, Dynamic Dilated Convolution Network (DDCN) \citep{nogueira2019dynamic}, SegNet, U-Net and FCN. Experimental results demonstrate that they have comparable F1-scores. DDCN achieves the best F1-score of 94.42\%, others attain 94.00$\sim$94.31\% with respect of F1-score. However, DDCN has the lowest detection efficiency with 1.02 min/ha, while other algorithms extract the area of each ha with only 15 seconds or so.

The second row of Table \ref{tab:tdl} summarizes the algorithms and collected examples for semantic segmentation-based tree delineation methods. In general, the semantic segmentation-based tree delineation methods are more efficient than sliding-window-based methods since they generate the detection results of the whole image at once. For example, \citet{brandt2020unexpectedly} extracts over 1.8 billion individual trees over a land area that covers 1.3 million km$^2$  in the West African Sahara, Sahel and sub-humid zone, with only 5\% of the annotated tree crowns are overlooked in the final results. Semantic segmentation-based tree delineation methods need to use an overlapping partition way for large-scale remote sensing image prediction into several image patches. In the meantime, every two adjacent patches in the whole image have an overlapping height (width) to make sure that corners are not missed by the algorithm (see Fig. \ref{fig:SSITC}). However, the performance of semantic segmentation-based tree delineation methods occur worse results for regions with tree crowns that appear to torch or overlap with each other, leading to segmenting some touching or overlapping tree crowns as only one tree crown. Besides that, the output of semantic segmentation-based tree delineation methods is a “confidence map” or a “probability map”, meaning the probability that a pixel belongs to the type of tree crown. These methods usually need a post-processing step to produce the final maps of an individual tree crown and segment overlapping tree crowns, for example, the local maximum detection \citep{freudenberg2019large}. Because of the manner of pixel-by-pixel classification, semantic segmentation-based tree delineation methods are more suitable for tree species mapping using semantic segmentation algorithms that do not need tree counting tasks \citep{morales2018automatic}.

%\subsubsection{Object detection based ITCD methods}

\section{Assessment for the accuracy of ITCD methods}
\label{sec:assessment}

Although \citet{yin2016assess} review the essential considerations and available techniques for evaluating detected individual tree locations and tree crown delineation maps using remote sensing data, we update some new ITCD evaluation metrics and observe some different conclusions on ITCD assessment according to our database of collected ITCD related publications. In addition, we further discuss tree crown counting assessment in this review. 

\subsection{Qualitative ITCD assessment}

As introduced in Sec. \ref{sec:evaluation}, qualitative and quantitative evaluation are two types of assessment for ITCD results.  On the one hand, qualitative evaluation is a straightforward visual comparison presenting the performance in a more direct way. There are various options for displaying the qualitative assessment, such as presenting the reference map and the resulting map in parallel \citep{brandtberg2002individual}, marking the delineated crown polygons \citep{holmgren2008species} or the detected tree top position \citep{chen2006isolating} on the reference map, etc. On the other hand, quantitative evaluation describes the accuracy with numbers and it is convenient to compare different methods by specific values.

\subsection{Quantitative tree crown detection assessment}

\label{sec:detass}

Different ITCD tasks have different evaluation metrics. For tree crown detection, we usually adopt True Positive (TP), False Positive (FP) and False Negative (FN) to describe the number of trees that are detected correctly,  the number of others that are detected as trees by model fault and the amount of ground-truth trees that are overlooked in detection results. According to these three indexes, we can calculate Precision, Recall, overall accuracy (OA) and F1-score. Precision and recall evaluate the algorithm’s capability of correctly detecting trees and the algorithm's capability of completely detecting ground-truth trees, respectively. OA and F1-score is to depict the overall results of the algorithm \citep{tao2020deep,miraki2021individual}:

\begin{equation}
    \begin{aligned}
        Precision &= \frac{TP}{TP + FP} \times 100\% \\
        Recall &= \frac{TP}{TP + FN}  \times 100\% \\ 
        OA &= \frac{Precision + Recall}{2} \times 100\% \\
        F1-score &= \frac{2 \times Precision \times Recall}{Precision + Recall} \times 100\% 
    \end{aligned}
    \label{eq1}
\end{equation}

Actually, Precision and Recall are also named User's Accuracy (UA)  and Producer's Accuracy (PA), respectively \citep{larsen2007single,pu2012comparative}, or Correctness and Completeness, respectively \citep{ozcan2017tree,duan2017novel}. In addition, Recall sometimes is considered as Detection Accuracy (DA) \citep{pouliot2002automated,xiao2020treetop}. Other researchers may adopt Omission Error (OE) and Commission Error (CE) to evaluate the results of tree crown detection \citep{wulder2002error,dong2018single}:

\begin{equation}
    \begin{aligned}
        OE &= \frac{FN}{TP + FN} \times 100\% \\
        CE &= \frac{FP}{TP + FP}  \times 100\% 
    \end{aligned}
    \label{eq2}
\end{equation}

Other overall tree detection accuracy metrics include Accuracy Index (AI) \citep{pouliot2005approaches,zhou2013mapping} and Matching score (M-score) \citep{larsen2011comparison}, which can be calculated as Eq. \ref{eq3}:

\begin{equation}
    \begin{aligned}
        AI &= \frac{TP - FP}{TP + FP}  \times 100\%  \\
        M-score &= \frac{TP}{TP + FN + FP} \times 100\%
    \end{aligned}
    \label{eq3}
\end{equation}

\citet{yin2016assess} conclude that DA is the most commonly used in individual tree crown detection assessment. However, according to our collected ITCD publications, F1-score has become the most popular overall tree crown detection accuracy metrics (F1-score, M-score, AI and DA). Specially, over half papers that adopt CNN classification and object detection based ITCD methods use F1-score to quantitatively describe the overall performance of their ITCD algorithms. Furthermore, mean Average Precision (mAP) start to gain more attention in the accuracy evaluation of tree crown detection \citep{weinstein2020deepforest,culman2020individual}. mAP both considers recall and precision into a single metric by calculating the area under the precision-recall curve resulting in a score ranging from 0 to 1, which is defined as the mean precision at a set of eleven equally spaced recall levels (from 0 to 1 with a step size of 0.1) by the Pascal VOC Challenge \citep{everingham2010pascal}. mAP can be formulated as Eq. \ref{eq4}:

\begin{equation}
    mAP = \frac{1}{11} \sum_{Recall \in \left\{ 0, 0.1, ..., 1 \right\}} {Precision\left(Recall\right)}
    \label{eq4}
\end{equation}

\subsection{Quantitative tree crown counting assessment}

As for tree crown counting task, it focuses on estimating the number of tree crowns and the assessment is to evaluate the difference between the tree number of estimated by algorithms ($N^{est}$) and the tree number of observed by experts ($N^{obs}$), which is considered as the ground-truth. Tree crown counting usually is regarded as a density regression problem. There are six commonly used statistical metrics in tree crown counting quantitative assessment, including Mean Bias (MB) \citep{wan2018improving}, Mean Absolute Error (MAE) \citep{osco2020convolutional}, Relative Mean Absolute Error (RMAE) \citep{khan2018remote},  R-Squared ($R^2$) \citep{crowther2015mapping}, Root Mean Squared Error (RMSE) \citep{yao2021tree} (or Mean Squared Error (MSE)) and Relative Root Means Squared Error (RRMSE) \citep{liu2021deep}. MB, MAE and RMSE (or MSE) calculate the average difference of the actual errors, absolute errors, squares of the errors, respectively. $R^2$ is the coefficient of determination, which estimates the correlation between the number of annotated and predicted trees. RMAE depicts the value of MAE accounts for the number of annotated trees, while RRMSE depicts the value of RMSE accounts for the total ground-truth number.   Eq. \ref{eq5} lists their detailed formulation:

\begin{equation}
    \begin{aligned}
        MB &= \frac{1}{S} \sum_{i=1}^{S} \left( N^{est}_i - N^{obs}_i \right) \\
        MAE &= \frac{1}{S} \sum_{i=1}^{S} \left | N^{est}_i - N^{obs}_i \right |  \\
        RMAE &= \frac{1}{S} \sum_{i=1}^{S} \left | \frac{N^{est}_i - N^{obs}_i}{N^{obs}_i}  \right | \times 100\% \\
        RMSE &= \sqrt{\frac{1}{S} \sum_{i=1}^{S} \left( N^{est}_i - N^{obs}_i \right)^2 }\\
        rRMSE &= \frac{\sqrt{\frac{1}{S} \sum_{i=1}^{S} \left( N^{est}_i - N^{obs}_i \right)^2 }}{\sum_{i=1}^S N^{obs}_i} \times 100\% \\
        R^2 &= 1 - \frac{\sum_{i=1}^{S} \left( N^{est}_i - N^{obs}_i \right)^2}{\sum_{i=1}^{S} \left( N^{obs}_i - \overline{N^{obs}} \right)^2},
    \end{aligned}
    \label{eq5}
\end{equation}

where $S$ represents the number of samples, such as pixel-level \citep{crowther2015mapping,rodriguez2021mapping} or image-level \citep{katoh2012improving,liu2021deep}, and the subscript $i$ represents the indexes of the sample. $\overline{N^{obs}}$ denotes then mean value of the tree number of observed by experts. Generally, lower MB, MAE, RMAE, RMSE, and RRMSE values, or higher $R^2$ denote better tree counting estimation.

\subsection{Quantitative tree crown delineation assessment}
\label{deaass}

Crown delineation segments the image into multiple parts, each of which is required to be one tree crown. To this end, the crown delineation performance can be assessed by segmentation accuracy evaluation. Similar to tree crown detection assessment, most of their evaluation metrics is also available to tree crown delineation assessment, while we adopt pixel-based rather than object-based (tree-based) evaluation. All metrics except mAP introduced in Sec. \ref{sec:detass} (Eq. \ref{eq1}-\ref{eq3}) can evaluate the performance of tree crown delineation. However, as it is difficult to assess the matching level between the delineation crown and the reference crown, the delineation result is more complex than the detection result. Some researches consider adopting matching rate to describe the performance of tree crown delineation according to the over- or under-segmentation rate of the segments \citep{ke2010active,jing2012individual}, to the overlapping rate of the segments \citep{wang2004individual,wagner2018individual, liu2015novel}, or to other self-defined segmentation criteria \citep{brandtberg1998automated,bunting2006delineation}. For example, 
\citet{liu2015novel} proposes producer's and user's accuracy to evaluate whether a tree sample is appropriately delineated by the ITCD model and whether a segment generated by the ITCD model appropriately represents the ground-truth tree segment. Recently, mean IoU (mIoU) has been adopted in tree crown delineation evaluation \citep{chadwick2020automatic,g2020tree}, which computes the tree crown area overlapped by manual delineation ($A^{ref}$) and generated delineation ($A^{est}$) (intersection area) divided by the sum of tree crown area from the manual delineation and generated delineation (union area). mIoU can be calculated by Eq. \ref{eq6}. Some papers present another similar matric \citep{tong2021delineation}, such as Jaccard score (J-score) \citep{dalponte2019individual} and Area Error Ratio (AER) \citep{wu2016individual}.

\begin{equation}
    mIoU = \frac{1}{S} \sum_{i=1}^S \frac{A^{ref}_i \cap A^{est}_i}{ A^{ref}_i \cup A^{est}_i}
    \label{eq6}
\end{equation}

\section{Discussions}
\label{sec:diss}

ITCD is of utmost importance for a comprehensive understanding of the ecological environment on both global and local scales. The meta-analysis presented is convenient to outline the past, current and potential future of ITCD for those who want to know about this specific domain. A thorough introduction of ITCD algorithms in this review may be interesting to them. In this section, we discuss three ITCD-related issues to further comprehend the ITCD domain.

\subsection{Multi-sensor data in ITCD domain}
\subsubsection{ Comparison between LiDAR data and optical remote sensing data in ITCD domain}

LiDAR is a critical data source for forestry inventory and ecological analysis \citep{calders2020terrestrial}, which has been increasingly adopted in individual tree crown detection and tree parameters estimation \citep{yin2019individual,wang2019situ,yrttimaa2020performance}, such as Diameter Breast Height (DBH), Leaf Area Index (LAI), AboveGround Biomass (AGB), etc. However, no matter the Terrestrial Laser Scanning (TLS) or Airborne Laser Scanning (ALS), most of the existing forestry inventory concentrates on region scales because of their difficulties and high cost for data collection. Although UAV equipped with laser scanning is a low-range-low-cost LiDAR system, its study area is even smaller than TLS and ALS systems. On the other hand, LiDAR measurements (such as ESA BIOMASS and  NASA GEDI) from satellites covering larger-scale areas do not satisfy research of the individual tree scale \citep{dubayah2020global} and they mainly focus on some tree parameters retrieval at a coarser scale. On the contrary, optical data captures the tree crown reflectance with more spectral, texture and semantic information using passive remote sensing instruments. This rich information is beneficial to represent the intrinsic features of vegetation and observe the conditions and status (such as disease). Furthermore, high-resolution and global optical data is much easier to acquire than LiDAR and has a large amount of storage data waiting for us to use over the last two decades. 
Actually, with high-resolution optical satellite data, we could soon map every tree on Earth \citep{hanan2020satellites}. The individual tree crown detection over a large area in West Africa \citep{brandt2020unexpectedly} suggests that it is possible to detect the location and size of every individual tree worldview according to existing optical satellite data. 
Although it is unable to provide 3D information, some researchers have explored the potential of side-view optical data (such as fisheye cameras and Google Street images) to better describe the tree trunk and branches, which will be introduced more in Sec. \ref{sec:multi}. 

\subsubsection{Multi-sensor fusion in ITCD domain}
Besides the revolution in algorithms, the prosperity of multi-sensor data also provides strong support for the development of ITCD domain. Multi-sensor data fusion plays a pivotal role in advancing ITCD in various environmental monitoring and remote sensing applications \citep{alonzo2014urban, quan2023tree, jones2010assessing}. The primary aim of multi-sensor data fusion is to integrate information from diverse sensors, such as optical, LiDAR, radar, and multispectral sensors, to improve the accuracy, completeness, and robustness of tree detection and delineation processes \citep{johansen2014mapping}. \citet{liu2017mapping} fuses LiDAR and hyperspectral data to map fifteen urban tree species which could provide both vertical and horizontal information and have shown great potential in improving tree species identification. \citet{alonzo2014urban} attempts to improve tree species classification results using crown-object level fusion of hyperspectral imagery and structural metrics extracted directly from the 3-D LiDAR point cloud. Multi-sensor data fusion in ITCD enables researchers to combine complementary data sources that capture different aspects of tree characteristics. For example, optical sensors provide valuable color and texture information, while LiDAR offers detailed 3D structural data. Radar sensors are proficient at penetrating vegetation, especially in adverse weather conditions. Multispectral sensors provide spectral signatures useful for discriminating between tree species. As a result of multi-sensor data fusion, ITCD algorithms benefit from enhanced spatial and spectral information. The fusion process aids in distinguishing between trees and other objects, accurately estimating tree height and crown diameter, and identifying changes in tree cover over time. The outcomes include more precise forest inventory, better forest management, and informed decision-making regarding biodiversity, carbon sequestration, and environmental conservation \citep{dalponte2014tree}.

\subsection{Difference between individual tree crown detection and delineation}
\label{sec:difference}

\begin{table}[t]
    \centering
    \caption{Detailed ITCD functions for different ITCD methods. $\checkmark$ and $\times$ denotes the ITCD method completely implement and fail to implement corresponding functions, respectively. $\checkmark ^ \sharp$ represents the ITCD method can implement corresponding functions through other pre-processing or post-processing procedures.}
    \resizebox{\textwidth}{!}{
    \begin{tabular}{cccccc}
    \hline
        \multicolumn{3}{c}{Method}  & \multirow{1}*{Detection} & \multirow{1}*{Delineation} & \multirow{1}*{Applications}  \\
        %Class & Division & Section \\
        \hline
       \multirow{2}*{Traditional image}   & \multicolumn{2}{c}{Local maximum filtering} & $\checkmark$ & $\times$ & $\checkmark$ \\
       \multirow{2}*{processing based}& \multicolumn{2}{c}{Image segmentation} & $\times$  &  $\checkmark$ & $\checkmark ^ \sharp$ \\
       \multirow{2}*{ITCD methods}& \multicolumn{2}{c}{Template maching} & $\checkmark ^ \sharp$ & $\times$ & $\checkmark ^ \sharp$ \\
       & \multicolumn{2}{c}{Image binarization} & $\checkmark ^ \sharp$ &  $\checkmark$ & $\checkmark ^ \sharp$  \\
       \hline
       Traditional machine learning  & \multicolumn{2}{c}{Patch-based} & $\checkmark ^ \sharp$ & $\times$ & $\checkmark ^ \sharp$ \\
       based ITCD methods & \multicolumn{2}{c}{Pixel-based} & $\checkmark ^ \sharp$  & $\checkmark$ & $\checkmark ^ \sharp$ \\
       \hline
     Deep learning based & \multicolumn{2}{c}{Semantic segmentation} & $\checkmark ^ \sharp$ & $\checkmark$ & $\checkmark ^ \sharp$ \\
       ITCD methods & \multirow{2}*{Object detection} & Mask R-CNN & $\checkmark$ & $\checkmark$ & $\checkmark$ \\
       & & Others & $\checkmark$ & $\times$ & $\checkmark$ \\
        \hline
    \end{tabular}
    }
    \label{tab:difference}
\end{table}

\begin{figure}[t]
    \centering
    \includegraphics[width=1.0\linewidth]{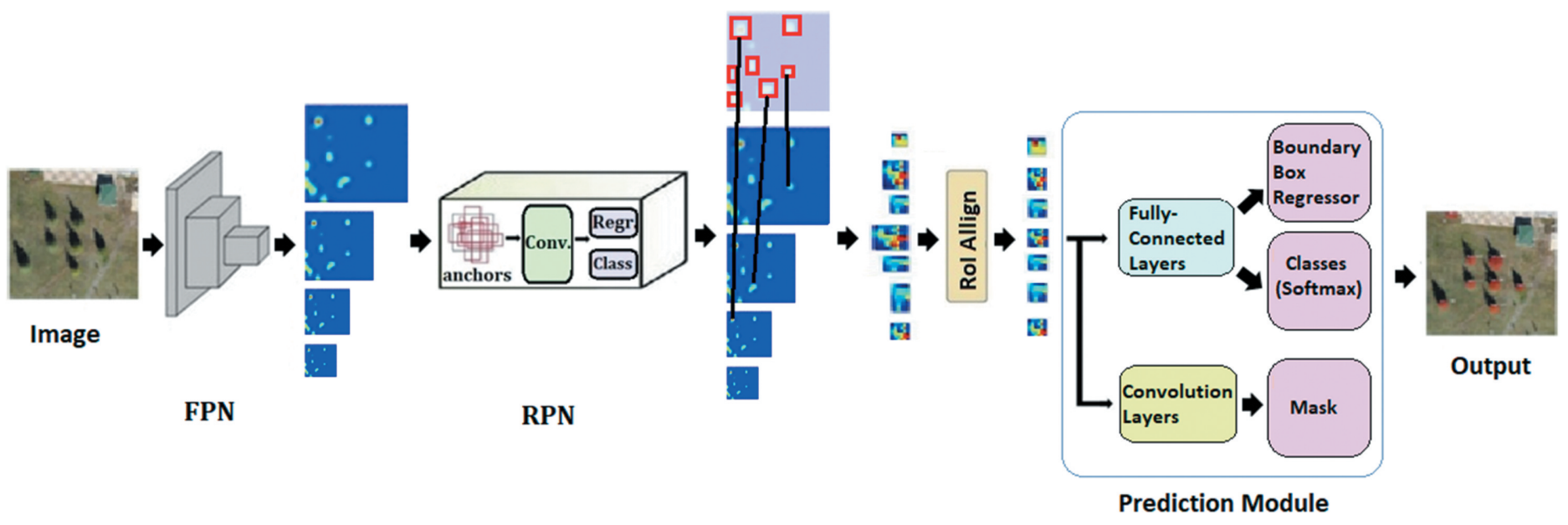}
    \caption{A typical ITCD example of object detection based ITCD method using Mask R-CNN proposed by \citet{ocer2020tree}.}
    \label{fig:mask}
\end{figure}

In this review, ITCD includes individual tree crown detection and individual tree crown delineation. Individual tree crown detection is mainly oriented to the location of individual trees, such as the center or the coordinates of four corners of the tree crown. Individual tree delineation mostly focuses on sketching the contour and shape of the tree crown or the area of tree crown canopy volume \citep{crowther2015mapping,rodriguez2021mapping}. Table \ref{tab:difference} lists detailed ITCD functions for different ITCD methods. In traditional image processing-based ITCD methods, local maximum filtering is the best at tree crown detection, while image segmentation is the best at tree crown delineation. Although image segmentation and image binarization can achieve tree crown counting, they probably require some post-processing steps. As for traditional machine learning-based ITCD methods, patch-based methods is similar to sliding-window-based ITCD methods, they usually need coordinates merging after image classification by pixel-based distance \citep{li2017deep} or Intersection-of-Union (IoU) metric \citep{zheng2020cross}. While pixel-based methods are similar to semantic segmentation-based ITCD methods, they are experts in individual tree crown delineation. On the contrary, since the common appearance of trees overlapping with each other, machine learning pixel-based and semantic segmentation-based ITCD methods require a post-processing procedure to produce the final location and contours of individual tree crowns, such as the local maximum detection \citep{freudenberg2019large,osco2020convolutional}. Most object detection methods can completely accomplish tree crown detection, while they are unable to conduct tree crown delineation except Mask R-CNN \citep{he2017mask}. As seen in Fig. \ref{fig:mask}, Mask R-CNN is an extension algorithm of Faster R-CNN \citep{ren2015faster}, combining both object detection and instance segmentation functions. To this end, Mask R-CNN is capable for all three ITCD tasks including tree crown detection and delineation.

\subsection{Comparison among different ITCD methods}

\begin{table}[t]
    \centering
    \caption{Qualitative comparison among different ITCD methods in annotation, efficiency and accuracy. +++ denotes the method performs best in this respect, while + denotes the method that performs worst in this respect.}
    %\resizebox{\textwidth}{!}{
    \begin{tabular}{ccccc}
    \hline
        \multicolumn{2}{c}{Method}  & \multirow{1}*{Annotation} & \multirow{1}*{Efficiency} & \multirow{1}*{Accuracy}  \\
        %Class & Division & Section \\
        \hline
       \multicolumn{2}{c}{Traditional image processing based ITCD methods} & +++ & +++ & +  \\
       \multicolumn{2}{c}{Traditional machine learning based ITCD methods} & ++ & + & +\\
       \multirow{2}*{Deep learning} & CNN classification &  ++ & + & +++\\
       \multirow{2}*{based ITCD methods}& Semantic segmentation & + & ++ & +++ \\
       & Object detection & ++ & ++ & +++ \\
       \hline
    \end{tabular}
    %}
    \label{tab:comparison}
\end{table}

Table \ref{tab:comparison} displays qualitative assessment for different ITCD methods in three aspects, including annotations, efficiency, and accuracy. We conduct in-depth discussions on these three aspects in this part.

\textbf{Annotation.} It is necessary and fundamental to conduct the annotation work in supervised learning. Traditional image processing-based ITCD methods have the least cost, and most of them are unsupervised learning methods and do not require any annotation work \citep{culvenor2002tida} except template matching \citep{larsen1998optimizing}. Annotation work of semantic segmentation-based ITCD methods is the most difficult and complex among all methods, since it is a pixel-level annotation and has to carefully outline all kinds of fine-grained tree crown shapes \citep{zhang2020identifying}. As for traditional machine learning-based ITCD methods and CNN classification methods, we not only have to annotate the samples of tree crowns but also have to annotate the samples of other land cover types, such as cropland, bare land, water, impervious area, etc \citep{mubin2019young}. As for object detection-based methods, we have to annotate the location of the four corners of a tree crown and generate a bounding box for each tree crown \citep{zheng2021growing}. Of course for Mask R-CNN, we further have to annotate the thorough shape of tree crowns to conduct tree segmentation \citep{lumnitz2021mapping}. The above three mentioned ITCD methods are of moderate difficulty and their annotation works are more difficult than traditional image processing-based ITCD methods while easier than semantic segmentation-based ITCD methods.

\textbf{Efficiency.} Algorithm efficiency is a crucial and key factor in ITCD applications, especially applied to large-scale study areas. Since most of them are unsupervised learning algorithms, traditional image processing-based ITCD methods cost most of the time in simple and basic image operations, usually with low computation complexity and fewer iteration times \citep{wulder2000local}. Traditional machine learning-based ITCD methods and CNN classification methods have the worst performance on algorithm and implementation efficiency, given that they require the time-consuming sliding window scheme to achieve the location and recognition of tree crowns \citep{wang2019automatic}. In addition, classifiers or neural network training and parameter tuning phases worsen their efficiency. Although semantic segmentation-based ITCD methods and object detection-based ITCD methods both have time-consuming neural network training work, they belong to end-to-end algorithms that allow the detection of several trees in the whole patch image \citep{weinstein2020deepforest}. To this end, these two algorithms are moderately efficient, higher than traditional machine learning-based ITCD methods while lower than machine learning-based ITCD methods.

\textbf{Accuracy.} ITCD accuracy is the most important evaluation to judge the ITCD algorithm whether successfully applied to practical tree inventory. In general, deep learning-based methods perform best in accuracy, with a high capacity of robustness and generalization \citep{martins2021semantic,flores2021automated}. In particular, deep learning-based ITCD methods achieve more convincing and satisfying results in complex areas \citep{weinstein2021remote}. Notably, semantic segmentation-based ITCD methods and object detection-based ITCD methods hold a little better performance than CNN classification methods. As for traditional image processing-based ITCD methods, their accuracy is generally the lowest among different algorithms and only has satisfied performance in simple areas or under specific parameters or specific regions. When the study sites turn to a varied topography, complicated environment or regions with overlapping crowns, the accuracy may incur a terrible deterioration. The accuracy of traditional machine learning-based ITCD methods is between that of deep learning-based ITCD methods and traditional image processing-based ITCD methods.

\begin{figure}[t]
    \centering
    \includegraphics[width=1.0\linewidth]{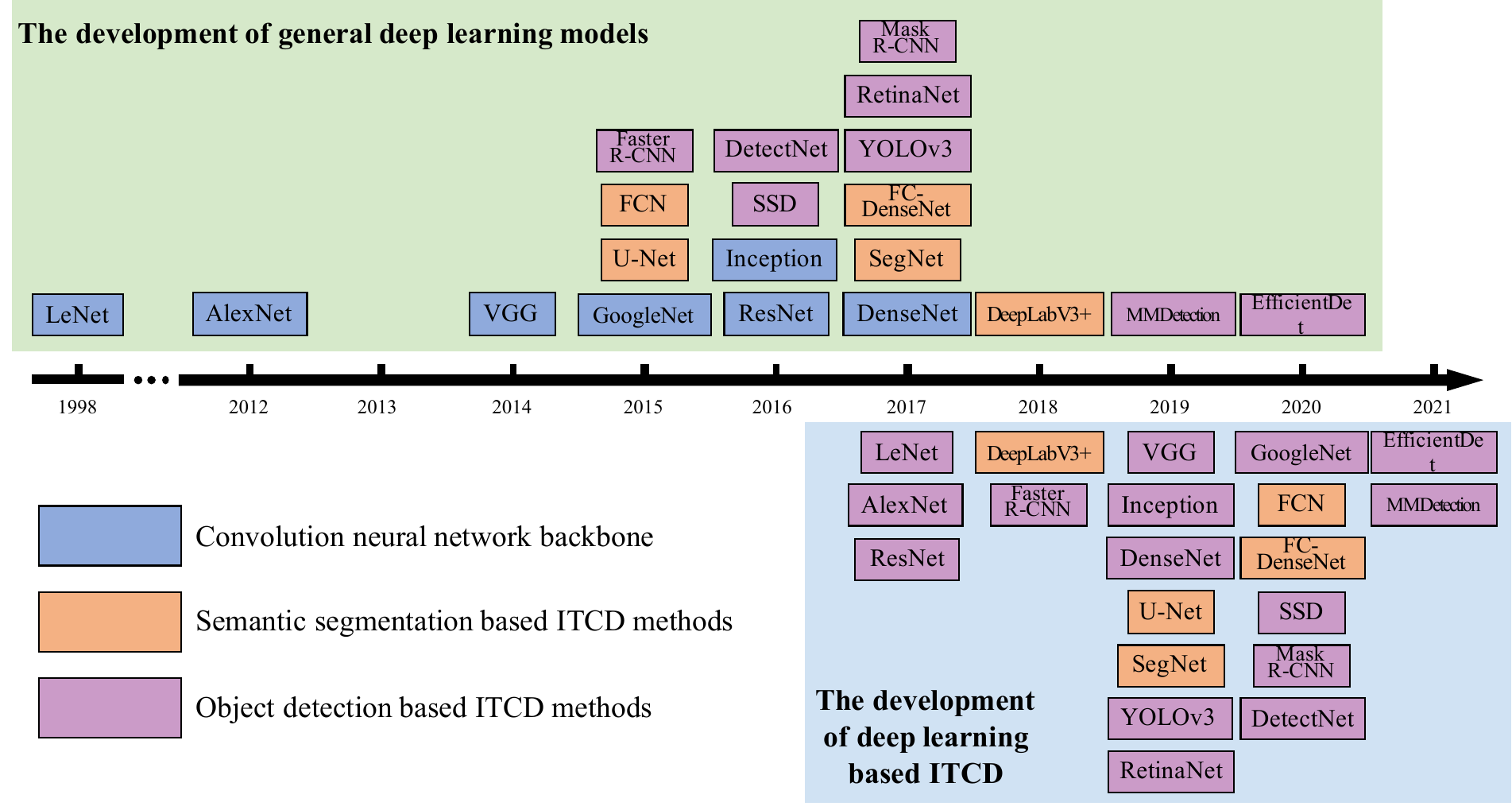}
    \caption{Comparisons between general deep learning models and their applications in ITCD domain.}
    \label{fig:dlcompare}
\end{figure}

\subsection{Comparisons between general deep learning models and their applications in ITCD domain}

In figure \ref{fig:dlcompare}, the top of the timeline shows the development of general deep learning architectures and the bottom of the timeline shows years of these deep learning models first used in ITCD domain. LeNet \citep{lecun1998gradient} and AlexNet \citep{krizhevsky2012imagenet} was proposed in 1998 and 2012, respectively, while they were applied in ITCD domain until 2017. After 2017, in general, novel deep learning models would be adopted in ITCD domain within three years. For example, Faster R-CNN \citep{ren2015faster}, Inception \citep{szegedy2016rethinking}, FC-DenseNet \citep{jegou2017one}, and Mask R-CNN \citep{he2017mask} are applied to ITCD domain after three years since they were proposed. YOLOv3 \citep{redmon2018yolov3}, RetinaNet \citep{lin2017focal} and MMDetection \citep{chen2019mmdetection} are adopted in ITCD applications after two years since they were firstly proposed. I costs only one year for ResNet \citep{he2016deep} and EfficientDet \citep{tan2020efficientdet} to be utilized in ITCD domain. Furthermore, DeepLavV3+ \citep{chen2018encoder} is applied to ITCD domain almost at the same time when it was proposed. To this end, the gap time between general deep learning models and their applications in ITCD domain is closer and closer. The progress of deep learning architectures plays a vital role in the development of deep learning-based ITCD methods.

According to different deep learning models, we can complete different ITCD tasks (see Table \ref{tab:difference}). General deep learning models can be directly applied to ITCD scenarios. However, some differences and modifications should be considered:

\begin{enumerate}
    \item Since the size of a remote sensing image is too large to be data input for a general deep learning model, we need to utilize an overlapping partition method for a large-scale remote sensing image into several sub-images in the inference phase (see Fig. \ref{fig:SSITC}). After that, we apply coordinates transformation and merge results of all sub-images to achieve the final ITCD results \citep{freudenberg2019large}.
    \item As for the design of deep learning architectures, we need to modify the sizes and the ratios of candidate anchor boxes in object detection-based ITCD methods, due to the size of tree crown is usually different from general objects \citep{zheng2021growing}. Furthermore, the number of channels in the first layer usually need to be modified because of multi-spectral bands for remote sensing images rather than only three bands for general images \citep{ampatzidis2019citrus}.
    \item Many deep learning based need to have some post-processing procedures. For example, the results of semantic segmentation-based ITCD methods is a “confidence map”, so that they usually require the local maximum detection or image segmentation to produce the final locations and contours of individual trees \citep{osco2020convolutional}. Some studies design a specific post-processing regularization to reduce the false positives \citep{zheng2019large} and improve the accuracy of ITCD model.
\end{enumerate}

\subsection{Dataset construction in ITCD for deep learning methods}

The bottleneck for deep learning applications regarding ITCD is the difficulty in collecting high-quality training samples. Here we select a list of representative papers in Table \ref{tab:dataset} to show how these papers address this bottleneck and achieve promising ITCD results. There are four important indexes in constructing a useful dataset. First, the data source is the base to build high-quality datasets. As can be seen, these selected publications use either aerial images or UAV images. Aerial images are mainly from high-resolution commercial satellites such as QuickBird, WorldView, etc., while most UAV images are collected by researchers themselves, with post-processing after the fly and before the construction. The second thing is the resolution. ITCD tasks require high-resolution images and these publications achieve resolutions below 1m. The highest resolution is 0.02m. The third thing is the image number. These publications all first manually annotate sufficient target samples for training, validating, and testing, regarding the application areas and the number of model parameters. Normally over 2,000 images for training are a necessity, which largely prevents models from overfitting. When the deep learning models have parameters over 100 million (i.e., transformers), the basic number of annotation images should increase. The fourth thing is the image size. Sometimes image size selection is a trade-off between accuracy and efficiency. To fully dig into the model potential, pre-surveying of the tree size is necessary for pre-defining the proper image size. Besides, we should note that only a few publications publish their annotated datasets for reproduction and re-creation. We should encourage researchers to publish their datasets to contribute to the whole community.

\begin{table}[t]
\caption{The statistics of representative publications using deep learning-based methods to collect data.}  
\label{tab:dataset}  
\linespread{1.4}
\resizebox{150mm}{43mm}{
\begin{tabular}{ccccccccc}
\hline
Publication & Source & Resolution & Species & \makecell[c]{Image\\number} & \makecell[c]{Image\\size} & \makecell[c]{Instance\\number} & Area & Availability \\
\hline
\citet{osco2020convolutional} & UAV & 0.129m & citrus & \makecell[c]{2,389\\for training} & 256 & 37,353 & 
70ha&$\times$ \\
\citet{li2017deep} & \makecell[c]{Aerial\\image} & 0.6m & palm & 20,000 & 17 & 100,000 & 7,500ha & $\checkmark$  \\
\citet{brandt2020unexpectedly} & \makecell[c]{Aerial\\image} & 0.5m & mixed & 334,000 & 256 & 89,899 & 5,000ha & $\times$ \\
\citet{zheng2020cross} & \makecell[c]{Aerial\\image} & 0.6m & palm & 40,000 & 17 & 431,101 & 7,429ha & $\checkmark$  \\
\citet{hao2021automated} & UAV & 0.02m & fir & 25,446 & 128 & 197,922 & 4ha & $\times$  \\
\citet{lassalle2022deep} & \makecell[c]{Aerial\\image} & 0.3m & mangrove & \makecell[c]{18,000\\for training} & 256 & NaN & 645ha & $\times$ \\
\citet{zheng2021growing}& UAV & 0.1m & palm & \makecell[c]{6,000\\for training} & 1024 & 363,877 & 3,700ha & $\checkmark$  \\
\citet{la2021multi} & UAV & 0.04m & mixed & \makecell[c]{14,000\\for training} & 128 & NaN & 30ha & $\times$  \\
\citet{safonova2019detection} & UAV &  0.04m & fir & \makecell[c]{3,520\\for training} & 150 & NaN & 10ha & $\times$  \\
\citet{zheng2023surveying} &  \makecell[c]{Aerial\\image} & 0.6m & coconut & \makecell[c]{2,000\\for training} & 512 & 136,500 & 1,475ha & $\times$  \\
\citet{albuquerque2022mapping} & UAV &  0.02m & mixed & \makecell[c]{903\\for training} & 1024 & NaN & 36ha & $\times$  \\
\citet{zhang2022multi}& UAV &  0.04m & mixed & \makecell[c]{1,603\\for training} & 1024 & NaN & 200ha & $\times$  \\
\hline
\end{tabular}
}
\end{table}
\subsection{Influencing factors of ITCD approaches} \label{sec:factors}

Due to the complexity of the different research subjects (e.g., mixed forests, specific tree species, etc.) with different attributes (e.g., areas, density, locations, etc.), it is desirable to design or select appropriate ITCD approaches addressing ITCD tasks under different scenarios. In this section, we discuss the multiple influencing factors of ITCD approaches.

\textbf{Tree Species.} We list the statistics of tree species in ITCD-related publications in Fig. \ref{fig:species}. According to our statistics, it turns out that researchers tend to use traditional image processing-based methods when the study subjects are mixed forests (63.72\%). 37.25\% of object detection-based methods in related publications are also used for mixed tree ITCD. There are barely traditional machine learning-based methods or semantic segmentation-based methods addressing ITCD tasks within mixed forests. Instead, traditional machine learning-based methods are adopted more on specific tree detection (i.e., urban trees, dead trees) whose distribution is sparse, and semantic segmentation-based methods are more utilized on trees (i.e., oil palm trees, banana trees) that gather together. This is not to say traditional machine learning-based methods or deep learning-based methods cannot handle mixed-tree scenarios. Mixed forests often involve a large number of trees with different species, and machine learning or deep learning-based methods are data-driven, so it is not cost-effective to annotate the mixed forests for training samples especially when the
application areas are small.

\textbf{Tree Density.} Tree density also influences the choice and performance of ITCD approaches. From a vertical comparison, even though 63.72\% of traditional image processing-based method publications handle mixed forest scenarios that own high tree density, their average performance (81.74\% @ F1-score) is lower than that of traditional image processing-based methods on scenarios with low tree density (84.69\% @ F1-score). It is the same with traditional machine learning-based methods and deep learning-based methods. From a horizontal comparison, deep learning-based methods have outstanding average performance compared with traditional machine learning-based methods and traditional image processing-based methods when the tree density is at the same level. However, it is important to note that when the density is sparse, it is not desirable to use semantic segmentation-based methods to map the individual trees. In a nutshell, the performance is negatively correlated with the tree density. Data-driven methods may have a better performance at the same tree density scenario compared with other methods, only if the training samples are sufficient.

\textbf{Forest Type and Structure.} We divide the forest into three types: boreal, temperate, and tropical, and two structures: forest and plantation. Boreal forests mainly consist of regular pine trees and spruce trees, which are suitable for traditional image processing-based methods and traditional machine learning-based methods because of the feature invariance. Temperate forests are complex for multiple tree species and high tree density. All methods' performance slightly drops in this scenario, so it is the application area that determines the choice of the utilized method. If the area is smaller than 1,000 ha, it is better to use traditional image processing-based methods or unsupervised traditional machine learning-based methods due to the efficiency. Otherwise, data-driven methods show their superiority. There are few ITCD studies focusing on tropical forests because of the severe overlapping of trees. For plantations like oil palm tree plantations and olive plantations, because of human management, there is barely any overlapping or tree shading. The proper distance between trees allows all kinds of methods could perform well. However, when the scenarios are complex (i.e., mixed with dead trees or growing trees), and the areas are large (i.e., larger than 1,000 ha), it is ideal to utilize deep learning-based methods.
\section{ITCD related Applications}
\label{sec:application}

In this section, we introduce some practical ITCD-related applications, such as tree parameter estimation, tree species classification, health monitoring, etc. Other ITCD applications includes wildfire potential estimation \citep{contreras2013developing},  plant heterogeneity \citep{hakkenberg2018modeling}, or wildlife protection \citep{owers2015remote,tochigi2018detection} and biodiversity research \citep{stokely2022experimental}, etc. Most of them first conduct individual tree crown detection and then further conduct other analyses on a single-tree scale. With the emergence of end-to-end deep learning techniques recently, we are able to achieve individual tree crown detection or delineation, along with individual tree species classification or health monitoring in the meantime.

\subsection{Tree species classification}
\label{sec:TSC}

Tree species classification is a valuable and important task in forest science, helping us to understand the role of tree's ecological functions \citep{fassnacht2016review}. Most previous individual tree species classification is a two-stage work, including individual tree crown detection or delineation, and then species recognition \citep{alonzo2014urban,liu2017mapping}. Researchers adopt common classifiers, such as RF \citep{kaminska2018species}, SVM \citep{dalponte2015semi}, etc. to classify collected features for each detected tree crown. Nowadays, CNN \citep{nezami2020tree} and 3D CNN \citep{mayra2021tree} have been more and more applied in tree species classification perform superior results under enormous input extracted features.
More recently, Mask R-CNN \citep{he2017mask} is able to achieve end-to-end both individual tree detection and delineation, even along with individual tree species classification (see the "Classes (Softmax)" in the  "Prediction Module" in Fig. \ref{fig:mask}) \citep{zhang2022multi}. That is, individual tree species classification is no more a two-stage workflow while becomes a more simple but effective one-stage workflow through object detection based ITCD methods \citep{rodriguez2021comparing}.

\subsection{Health monitoring}

Multi-spectral information from optical remote sensing data, coupled with machine learning or deep learning techniques, play a considerable role in tree health monitoring, including disease surveillance, growing status observation, tree mortality mapping, etc. Similar to tree species classification (see Sec. \ref{sec:TSC}), previous popular individual tree's health monitoring is a two-stage work \citep{einzmann2021early}, while as deep learning-based ITCD methods emerge, existing tree's health assessment becomes a one-stage framework, both accomplishing individual tree crown detection and their status observation \citep{johansen2020mapping,zheng2021growing}. Compared to LiDAR data, deep learning may perform better on multi-spectral optical remote sensing data because of its rich semantic and texture information, which is also beneficial for capturing vegetation's intrinsic features. 
As a matter of fact, employing comprehensive health monitoring, especially for economical tree species, is beneficial to improve their productivity or yield, and further increase the economic effect \citep{alam2012improving}.

\subsection{Tree parameters estimation}

Tree parameters are quite vital biophysical representation that influences water, energy and carbon exchanges between the atmosphere and forest ecosystems. As aerial LiDAR and terrestrial LiDAR are able to acquire 3D point returns, most existing studies adopt them to estimate most of tree-related parameters, including first-order properties (such as height, crown diameters, etc.) \citep{yin2019individual} and second-order properties (such as basal area, aboveground biomass, etc) \citep{goldbergs2018hierarchical}. On the other hand, optical remote sensing data is also widely applied in studies that link tree parameters from the field to observations through the sensitivity of optical reflectance to canopy structure variations. Optical remote sensing provides great potential for tree parameter estimation at a larger scale than LiDAR data \citep{addink2007importance,karna2015integration}. For example, \citep{brandt2020unexpectedly} analyze the canopy cover, tree density and tree crown size over 1.3 million km$^2$ in West Africa, after detecting trees by semantic segmentation based ITCD method. However, optical remote sensing data is poor in height-related parameters \citep{song2007estimating}, which means that combining LiDAR or adopting side-view remote sensing data may address this issue. Furthermore, end-to-end forest attribute retrieving in the one-stage framework for individual trees still needs to be exploited and developed in the future.

\subsection{Multi-temporal change analysis}

Multi-temporal remote sensing data is able to not only conduct individual tree crown detection, but also explore the changes of individual crown diameters, canopy cover, growth process etc. evaluating the variants of ecological restoration \citep{gartner2014object}, carbon stock \citep{turner2019approach}, or the impacts of tree species competition \citep{ma2018quantifying} and natural disasters \citep{vastaranta2012mapping}. For example, \citet{blackman2020detecting} detects long-term urban forest cover change between 1938 and 2019 using high-resolution aerial images and LiDAR data, along with examining the impacts of typhoons and tree disease. Also, multi-temporal data contributes to better accomplishing individual tree crown detection and ITCD-related applications through seasonal spectral and texture variations \citep{papecs2013seasonal}. However, as the same as tree parameters estimation, and existing single tree-level change analyses are all two-stage works. Following the development of  recurrent neural networks, we believe that coupling with semantic segmentation and object detection-based ITCD methods with deep learning-based time series analysis may achieve real-time, high-accuracy and end-to-end single tree-level change analysis using multi-temporal remote sensing data.

\section{Prospects}
\label{sec:pros}

Based above literature analysis, methodology review and in-depth discussion, ITCD-related prospects emerged from this attempt, which concerns past, current and future trends. These prospects are introduced in this section. 

\subsection{ITCD using multi-source and multi-view remote sensing data}
\label{sec:multi}

Some ITCD researchers combine optical remote sensing data with other remote sensing data to extract high-dimension features, such as point clouds from LiDAR \citep{dalponte2015semi}, Digital Topographic Model (DTM) \citep{windrim2020tree}, Digital Surface Model \citep{chen2021assessment} or GIS data \citep{aval2018detection}. Meanwhile, some papers simultaneously adopt satellite and UAV images to achieve ITCD. For instance, \citet{selvaraj2020detection} firstly utilizes multispectral satellite images (WorldView-2, Panet and Sentinel-2) to extract banana plantation regions, and then use UAV images to precisely locate each banana plant and recognize its health condition. 

However, existing studies have not exploited the potential of fully fusing multi-source remote sensing data. For example, most of the UAV images are unable to describe abundant spectral information because they only have three bands (red, green and blue) rather than multispectral images photographed by the high expense of multispectral-based cameras. If we make full use of high-spatial-resolution UAV images and high-spectral-resolution satellite images \citep{alvarez2020can}, it is considerably beneficial to precisely recognize tree crowns and classify fine-grained tree species or growing status with high-spectral-spatial-resolution remote sensing data. On the other hand, remote sensing data acquired from a vertical view can not perfectly extract the overlapping tree crowns or those sheltered from higher mature trees with larger crowns. If it is available for remote sensing data from the side view, we are capable of recognizing those trees that are easily missed from vertical-view remote sensing data. Some researchers have attempted to detect and delineate individual tree crowns using side-view remote sensing data \citep{cheng2021cherry}.  We believe that integrating vertical-view and side-view remote sensing data is able to achieve better ITCD performance. In addition, combining Synthetic Aperture Radar (SAR) \citep{magnard2016single} or Interferometric SAR (InSAR) \citep{yazdani2020comparison} with LiDAR or optical remote sensing data also seems promising for ITCD domain, and we may pay more attention in exploiting other observation platforms with optical remote sensing data in the future.

\subsection{Fine-grained individual tree species or growing status classification}

Fine-grained individual tree classification includes fine-grained tree species classification and fine-grained growing status classification. The former is significant for understanding the distribution of forest species and protecting biodiversity. The latter not only observes damaged or diseased trees to prevent their proliferation but also is conducive to estimating the yield and increasing benefits for some economic trees. To this end, fine-grained tree classification has an extremely high value in both ecology and economy. Most of the existing individual tree species studies focus on small areas (smaller than 1000ha) \citep{fassnacht2016review}. Although \citet{slik2015estimate} estimate the number of tree species in the tropical area, they have not mapped the distribution of fine-grained tree species. As for growing status observation, most of them only classify into two statuses: healthy tree and unhealthy tree \citep{selvaraj2020detection}. In contrast, few studies are devoted to multi-classes growing status classification \citep{johansen2020mapping}. It is high-demanded for plantations (such as oil palm, olive, etc.) to monitor more fine-grained healthy conditions, such as specific diseases. Until now, individual tree classification work is a two-stage scheme, firstly detecting or delineating individual tree crowns and then completing species or growing status recognition. Furthermore, fine-grained classification requires recognizing the slight difference between similar classes, which may need high-spatial and high-spectral resolution remote sensing images. The data-fusing approaches mentioned in Sec. \ref{sec:multi} will be an effective solution.

\subsection{Large-scale ITCD in spatial big data era}

Undoubtedly, we are in the big data era at present and in the future. Massive remote sensing images acquired by satellites, aerial planes, UAVs and even mobile phones create the spatial big data era. With these earth observation data, we have the opportunity to achieve large-scale ITCD, and deeply comprehend global tree resources. Most of the study area in existing ITCD research is smaller than 1000km$^2$ except \citet{brandt2020unexpectedly}. They extract over 1.8 billion individual trees over a land area that spans 1.3 million km$^2$ in the West African Sahara, Sahel and sub-humid zone, with only 5\% of the labeled trees being overlooked in the final results. Despite the fact that \citet{crowther2015mapping} roughly estimate there are 3.04 trillion trees around the world, they only approximately map the global tree distribution and tree density. There are two major challenges in large-scale ITCD work. The first one is the capacity of model generalization. As we have to prepare multi-temporal, multi-source and multi-regional remote sensing data to conduct large-scale ITCD \citep{wu2020cross}, developing a more transferable, robust and general model is a powerful foundation using advanced algorithms, such as domain generalization, domain adaptation and transfer learning. Another challenge is the capacity of computation performance to support the efficiency of ITCD in large-scale areas. At present, some studies adopt high-performance computation platforms (such as FPGA and GPU) to accelerate ITCD algorithms \citep{jiang2017papaya}. However, global, continental or national-level ITCD research has not been completed in higher-performance computing platforms such as supercomputers, which may be a potential general platform for processing global observation issues.

\section{Conclusions}
\label{sec:concl}

Individual Tree Crown Detection (ITCD) using high-resolution optical remote sensing data is essential for forestry inventory and ecological analysis in an automated way.  In this review, a comprehensive overview of ITCD-related research is introduced. First, we conduct an investigation of scientific peer-reviewed journal papers over 20 years building an available database and carrying out a meta-analysis. Second, intriguing and thorough ITCD methods are presented that depict the trend and development of past years relating to this specific domain. We classify ITCD methods into three types, including traditional image processing-based ITCD method (such as local maximum filtering, image segmentation, etc), traditional machine learning-based ITCD method (such as random forest, decision tree, etc.) and deep learning-based ITCD method. In addition, we also categorize deep learning-based ITCD methods into three types (i.e., CNN classification, semantic segmentation and object detection) and discuss their pros and cons. With the current pace at the methodology of ITCD research is conducted, such information is rather essential and truly valuable. In addition, we discuss three ITCD-related topics to further comprehend the ITCD domain, such as comparisons between LiDAR data and optical remote sensing data, comparisons among different algorithms and different ITCD tasks. Finally, some ITCD-related applications and a few existing and emerging topics are presented, and we promise the significance and prosperity of ITCD in the future.

\section*{Acknowledgements}

This work is partially supported by National Key Research and Development Program of China (2017YFA0604500), and National Natural Science Foundation of China (U1839206).

%\renewcommand{\baselinestretch}{0.15}
%\scriptsize
\bibliography{main}

\end{document}